\documentclass[final,leqno,onefignum,onetabnum]{siamltexmm}
\usepackage[utf8]{inputenc}
\usepackage{amssymb,amsmath,dsfont}
\usepackage{stmaryrd}
\usepackage{epsfig}
\usepackage{subfigure}
\usepackage{breqn}
\DeclareMathOperator{\argmin}{\text{argmin}}
\usepackage{mathtools}

\DeclarePairedDelimiter\floor{\lfloor}{\rfloor}

\usepackage{algorithmic}
\usepackage{algorithm}

\title{PSF field learning based on Optimal Transport distances} 

\author{F. Ngolè$^1$ and J.-L. Starck$^1$}

\begin{document}
\maketitle
\newcommand{\slugmaster}{%
\slugger{siims}{xxxx}{xx}{x}{x--x}}

\begin{abstract}
\underline{Context}: in astronomy, observing large fractions of the sky within a reasonable amount of time implies using large
field-of-view (fov) optical instruments that typically have a spatially varying Point Spread Function (PSF). Depending on the scientific goals, galaxies images need to be corrected for the PSF whereas no direct measurement of the PSF is available. 
  
\underline{Aims}:  given a set of PSFs observed at random locations, we want to estimate the PSFs at galaxies locations for shapes measurements correction.	

\underline{Contributions}: we propose an interpolation framework based on Sliced Optimal Transport. A non-linear dimension reduction is first performed based on local pairwise approximated Wasserstein distances. A low dimensional representation of the unknown PSFs is then estimated, which in turn is used to derive representations of those PSFs in the Wasserstein metric. Finally, the interpolated PSFs are calculated as approximated Wasserstein barycenters.

\underline{Results}: the proposed method was tested on simulated monochromatic PSFs of the Euclid space mission telescope (to be launched in 2020). It achieves a remarkable accuracy in terms of pixels values and shape compared to standard methods such as Inverse Distance Weighting or Radial Basis Function based interpolation methods. 

\end{abstract}


\begin{AMS}{Numerical methods; Algorithms; Mathematical programming, optimization and variational techniques}\end{AMS}

\pagestyle{myheadings}
\thispagestyle{plain}
\markboth{PSF field interpolation}{Optimal Transport}

\section{Introduction}
\subsection{Context}
The Point Spread Function (PSF) modeling is an essential step in a wide range of imaging applications in science, as the PSF translates instrument related distortions that have to be taken into account for quantitative analysis of images. 
An exciting example is found in cosmology. Indeed, it is possible to get precise knowledge of the Universe geometry and dynamic at large scales from observed galaxies shapes, provided that those shapes are corrected from the PSF. This is precisely one of the main goals of the ESA's Euclid mission \cite{Eucl1}, to be launched in 2020. 
For wide field-of-view (fov) instruments such as the Euclid telescope, the PSF varies substantially across the focal plane. Assuming that the PSF can be modeled locally as a convolution kernel, we use "PSF field" to refer to a continuous function mapping each point in the instrument focal plane surface to the corresponding convolution kernel. In astronomy, unresolved sources images like isolated stars give a measurement of those kernels. In \cite{fng}, a method for restoring a PSF field at the available degraded measurements locations was proposed. This work rather focuses on interpolating a PSF field at arbitrary locations from a set of perfectly known PSFs spread out across the fov. This problem has driven a lot of attention within the astronomers community over the last decade\cite{gentile_2013}, especially due to the strengthening of accuracy requirements in the recent and future spatial surveys. Indeed future missions such as Euclid will require an extraordinary accuracy on the galaxies shape measurements, which implies to use the most advanced mathematical tools, if we want to meet the requirements.   

\subsection{State-of-the-art}
\label{state}
Most of the PSF interpolation methods fall into one of the two following categories: 
\begin{itemize}
\item[•] optical model based methods which derives the PSF from a physical model of the instrument itself;
\item[•] data driven methods which estimate the PSFs at given locations in the fov based on local measurements extracted from real images. 
\end{itemize} 
In the first category, we can mention the work in \cite{jar_2008}. The authors built a parametric model of the PSF shape and size by modeling individually the main physical causes which make the PSF variable; they found that this model is able to reproduce a substantial part of the PSF anisotropy and size for the ground based telescope Blanco telescope. In \cite{stab}, the authors simulate the PSF temporal variations due to thermal drift, jitter and structural vibrations, using the method of ray tracing, for the SNAP telescope. However, these methods require making assumptions on the instrument physic, and not all the phenomena at play in the image forming process can be satisfactorily modeled. Thus, we only consider model-free methods for comparisons in the numerical experiments.

The data driven approaches rely on the fact that in practice, unresolved stars images in the observed field give a measurement of the PSFs at those stars locations. One recurrent scheme consists in first expanding the observed unresolved stars over some given analytic function basis. This gives a more compact representation of the data (and potentially allows denoising). The PSF field is then obtained through a 2D polynomial fitting typically, of the representation coefficients for each element of the chosen basis. In \cite{romano}, this scheme is applied to model the PSF of the Large Binocular Cameras, using the Shapelets basis (\cite{shap1,shap2}). However, it might not be possible to capture all the PSF structures through a finite expansion over an analytic function basis. One can instead learn the representation basis from the data themselves. For example in \cite{jee_2007}, a principal components analysis (PCA) is performed over hundreds archival images of stars from the Hubble Space Telescope (HST) Advanced Camera for Surveys (ACS), from different exposures. The optimal representation basis is then chosen as the first principal components and the representation coefficients are fitted in each exposure separately with a bivariate polynomial. In \cite{jarv_2004}, the PSF modeling also relies on a PCA, not of the images themselves, but instead of a set of features (for example a shape parameter). In the two aforementioned methods, fitting a global polynomial model can yield an oversmoothing of the PSF field, or instead a Runge phenomenon, if the polynomial's degree is too high \cite{epperson1987runge}.  In \cite{gentile_2013}, the PSF modeling is recast as a spatial interpolation problem. In general statistics, the spatial interpolation methods aim at interpolating a field exploiting the spatial correlation of the data. Some of these methods, for example the Kriging, are presented and applied to PSF interpolation in the previously mentioned work. We consider the two best performing approaches tested in \cite{gentile_2013} for comparisons in the numerical experiments.
More recently a PSF interpolation scheme has been proposed in \cite{interp_cs} based on compressive sampling (CS) ideas. The method aims at computing the PSF ellipticity parameters at every points over an uniform grid using the available randomly distributed measurements. This yields an ill-posed inverse problem which is regularized by assuming that the shape parameters constitute a compressible field in frequency domain. However, this approach has a narrow scope of applications since the "full" PSF itself is not estimated.       

Let finally mention a somehow hybrid approach. Indeed in \cite{piot}, the authors propose to build an accurate polynomial model of the PSF of the "Pi of the Sky" telescope from detailed laboratory measurements of optical PSF, pixels sensitivity and pixels response. This model is then tuned to real sky data.

In this paper, we propose a non-parametric interpolation method that relies on geometric tools. We propose an intuitive framework for learning a PSFs set underlying geometry using Optimal Transport distances. In Section \ref{ot_int}, we introduce and motivate the mathematical framework; we describe the proposed method and algorithm in Section \ref{train_meth}; we present some numerical experiments and discuss the results in Section \ref{num_res}; the interpolation algorithm parameters and some numerical considerations as discussed in Section \ref{practical}.

\subsection{Notations}
We adopt the following notation conventions:
\begin{itemize}
\item we use bold low case letters for vectors;
\item we use bold capital case letters for matrices;
\item the vectors are treated as column vectors unless explicitly mentioned otherwise. 
\end{itemize}
Each PSF is treated either as a matrix $\mathbf{X} = (x_{ij})_{1\leq i\leq n \atop 1\leq j\leq n}$ or as a vector in $\mathbf{R}^{n^2}$, $\mathbf{x} = (x_{k})_{1\leq k\leq n^2}$. For $N \in \mathbb{N}$, we denote $\Sigma_N$ the set of permutations of $\llbracket 1,N\rrbracket$.
We denote $\mathcal{P}(\mathbb{R}^d)$ the set of discrete probability measures over $\mathbb{R}^d$.
We denote $\mathds{1}_n$ the column vector of $n$ ones.
We denote $\mathbf{I}_p$ the identity matrix of size $p\times p$.

\section{Introduction to Optimal Transport and motivation}
\label{ot_int}
\subsection{Motivation}
\begin{figure}
\begin{center}
\includegraphics[scale=0.5]{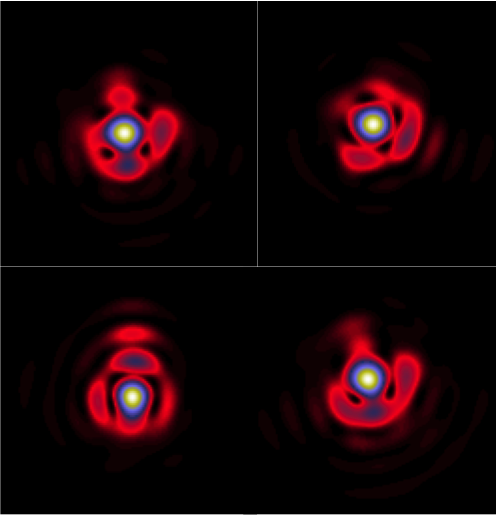}
\end{center}
\caption{Simulated Euclid telescope PSFs at different locations in the FOV; source: Koryo Okumura et al.}
\label{psf_var}
\end{figure}

In this work, we consider a field of monochromatic PSFs. The only factor of variability is the unresolved object's position in the FOV. Formally, the optical PSF $h[{\mathbf{p}}]$ at a location $\mathbf{p}$ in the FOV takes the following form \cite{goodman2005introduction}:
\begin{equation}
h[{\mathbf{p}}](u,v) \sim |\int\int_{\mathbb{R}^2}P[\mathbf{p}](x,y)e^{j\frac{2\pi}{\lambda}W[\mathbf{p}](x,y)}e^{-j\frac{2\pi}{\lambda c}(x u+y v)} dxdy|^2
\label{ampl_psf}
\end{equation}
where $c$ is a constant, $\lambda$ the light wavelength, $P[\mathbf{p}]$ is the space-dependent pupil function and the phase term $W[\mathbf{p}]$ represents the optical aberrations effect at $\mathbf{p}$.
This translates into a non-linear variation of the PSF intensity distribution that is illustrated in Fig.~\ref{psf_var}.
%

\begin{figure}
\begin{center}
\includegraphics[scale=0.5]{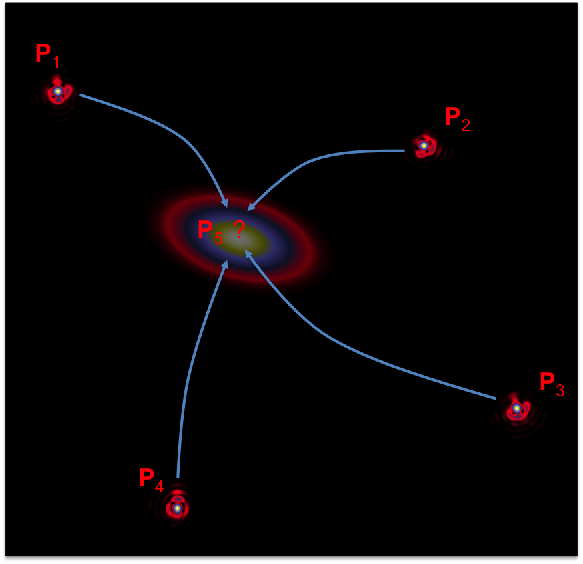}
\end{center}
\caption{PSF interpolation problem}
\label{psf_interp}
\end{figure}

It can be interpreted as the curvature in $\mathbb{R}^{n^2}$ of the PSFs manifold, occurring between distant PSFs in the FOV.
As Fig.~\ref{psf_interp} shows, we want to combine available PSFs measurements given by unresolved objects images to estimate the PSF at galaxies locations in the FOV.


Optimal Transport (OT) appears to be suitable to tackle this problem. Indeed, it gives a way of measuring distances between PSFs which accounts for the aforementioned curvature. 

Besides, it comes with a recipe for computing geodesics (see Section \ref{not}) which is interesting for interpolation purpose. Moreover, it is classical to assume that a PSF has an unitary $l_1$ norm. This implies a constraint of mass conservation in the interpolation which is naturally integrated in the OT framework. 

In the next section, we precise the OT tools and concepts we use.

\subsection{General notions}
\label{not}
\subsubsection{Sliced Wassertein Distance}
We consider two points matrices $\mathbf{X} = [\mathbf{x}_1,\dots,\mathbf{x}_N]$ and $\mathbf{Y} = [\mathbf{y}_1,\dots,\mathbf{y}_N]$ in $\mathbb{R}^{d\times N}$, for some integers $d$ and $N$. By abuse of language, we do not distinguish these matrices and the point clouds in $\mathbb{R}^d$ they represent, where there is no risk of confusion. We define the distributions $\mu_{\mathbf{X}} = \sum_{i=1}^{N}\frac{1}{N}\delta_{\mathbf{x}_{i}}$ and $\mu_{\mathbf{Y}} = \sum_{i=1}^{N}\frac{1}{N}\delta_{\mathbf{y}_{i}}$, where for $\mathbf{x}$ and $\mathbf{y}$ in $\mathbb{R}^d$, $\delta_{\mathbf{x}}(\mathbf{y}) = 1$ if $\mathbf{x}=\mathbf{y}$ and $\delta_{\mathbf{x}}(\mathbf{y}) = 0$ otherwise.
The optimal assignment problem between $\mu_{\mathbf{X}}$ and $\mu_{\mathbf{Y}}$ consists in finding a permutation $\sigma \in \Sigma_N$ which minimizes
\begin{equation}
\sum_{i=1}^N \|\mathbf{x}_i-\mathbf{y}_{\sigma(i)}\|^p,
\label{opt_assign}	
\end{equation}
where $\|.\|$ is a norm in $\mathbb{R}^d$ and $p \geq 1$. Problem \ref{opt_assign} can be rewritten as a particular instance of the Monge-Kantorovich mass transportation problem\cite{kant}.   
We note $\sigma^*$ a minimizer of Problem \ref{opt_assign}. It has been shown that $W_p(\mu_{\mathbf{X}},\mu_{\mathbf{Y}}) = (\sum_{i=1}^N \|\mathbf{x}_i-\mathbf{y}_{\sigma^*(i)}\|_p^p)^{\frac{1}{p}}$ defines a distance on $\mathcal{P}(\mathbb{R}^d)$ which is the so-called Wasserstein Distance $p$ of the two distributions. The norm $\|.\|_p$ is often referred to as the \underline{ground metric} and is an important parameter of the Transport Distances in practical applications. Examples of theoretical and computational applications of optimal transport can be found in \cite{not_prox}.

\paragraph{Displacement interpolation}

Let consider a point $\mathbf{x}_i$ in the first point cloud and the assigned point $\mathbf{y}_{\sigma^*(i)}$ in the second point cloud. Let $\gamma_i: [0,1] \mapsto \mathbb{R}^d$ a curve verifying $\gamma_i(0) = \mathbf{x}_i$ and $\gamma_i(1) = \mathbf{y}_{\sigma^*(i)}$. The action on a material particle moving from $\mathbf{x}_i$ to $\mathbf{y}_{\sigma^*(i)}$  along $\gamma_i$ between time $s$ and time $t$ might be defined as
\begin{equation}
\mathcal{A}_p^{s,t}(\gamma_i) = \int_s^t \|\dot{\gamma_i}(\tau)\|^p d \tau,
\label{action}
\end{equation} 
where $\dot{\gamma_i}(\tau)$ is the particle velocity at time $\tau$.
We define the set $\Gamma_i = \{\gamma: [0,1] \mapsto \mathbb{R}^d,  \gamma(0) = \mathbf{x}_i, \gamma(1) = \mathbf{y}_{\sigma^*(i)}\}$. 
An action minimizing path between $\mathbf{x}_i$ 	and $\mathbf{y}_{\sigma^*(i)}$ is a solution of 
\begin{equation}
\underset{\gamma \in \Gamma_i}{{{\text{min}}}} \mathcal{A}_p^{s,t}(\gamma).
\end{equation}

Interestingly, it has been shown that a geodesic path between the distributions $\mu_{\mathbf{X}}$ and $\mu_{\mathbf{Y}}$ in $\mathcal{P}(\mathbb{R}^d)$ equipped with the Wassertein distance can be computed by advecting particles along action minimizing paths between pairs of assigned points in the two point clouds \cite{vill_2}. 
Concretely, let $\gamma_i*$ denotes an action minimizing path between $\mathbf{x}_i$ and $\mathbf{y}_{\sigma^*(i)}$. For $t \in [0,1]$, we define the discrete distribution
\begin{equation}
\mu_t = \sum_{i=1}^N \frac{1}{N}\delta_{\gamma_i*(t)}.
\end{equation}

Then the parametric curve
\begin{equation}
\Gamma_{XY} : [0,1] \mapsto \mathcal{P}(\mathbb{R}^d), t \rightarrow \mu_t
\label{disp_interp}
\end{equation}
is a geodesic path in $\mathcal{P}(\mathbb{R}^d)$ between the distributions $\mu_{\mathbf{X}}$ and $\mu_{\mathbf{Y}}$. This realizes a \textit{displacement interpolation} of the two considered distributions; this powerful notion has been introduced in \cite{McCann1997153}. For a given $t$, the distribution $\mu_t$ can be seen as the barycenter of the two end distributions with the barycentric weights $1-t$ and $t$ respectively in the Wasserstein metric.

For $d = 1$, i.e. in the 1D case, $\sigma^*$ is known in closed form: if we consider two permutations $\sigma_X$ and $\sigma_Y$ in $\Sigma_N$ verifying
\begin{eqnarray}
\label{sorting}
\mathbf{x}_{\sigma_X(1)} \leq \dots \leq \mathbf{x}_{\sigma_X(N)} \\
\mathbf{x}_{\sigma_Y(1)} \leq \dots \leq \mathbf{x}_{\sigma_Y(N)},
\end{eqnarray}  
then  $\sigma^* = \sigma_Y\circ\sigma_X^{-1}$, where $\sigma_X^{-1}$ is a permutation verifying $\sigma_X^{-1}\circ\sigma_X = \text{Id}$.
For $d>1$, there is no known closed-form expression for $\sigma^*$.  Various methods have been proposed for solving Problem \ref{opt_assign}; in particular, it can be recasted as a linear program. Yet, the fastest known algorithms have a running time of $O(N^{2.5}\text{log}(N))$ \cite{opt_assign}, which is prohibitive for large scale image processing applications. This has motivated the introduction of the sliced Wasserstein Distance \cite{sliced_wass,sliced_bonneel}, which consists in the sum of 1D Wassertein distances of the projected point clouds:
\begin{equation}
SW_p(\mu_{\mathbf{X}},\mu_{\mathbf{Y}})^p = \int_{\mathbb{S}^{d-1}} W_p(\mu_{\mathbf{X}_{\mathbf{u}}},\mu_{\mathbf{Y}_{\mathbf{u}}})^p d\mathbf{u},
\end{equation}
where $p\geq 1$, $\mathbb{S}^{d-1}= \{\mathbf{x}\in \mathbb{R}^d / \|\mathbf{x}\| = 1\}$, $\mathbf{X}_{\mathbf{u}} = \{\mathbf{u}^T\mathbf{x}_i, i = 1...N \} \subset \mathbb{R}^N$ and $\mathbf{Y}_{\mathbf{u}}$ is similarly defined.
\subsubsection{Sliced Wassertein Barycenter}
\label{wass_bar_section}
We consider a family $\{\mathbf{X}_1,\dots,\mathbf{X}_K\}$ of point clouds in $\mathbb{R}^d$: for $k \in \llbracket 1,K \rrbracket$, $\mathbf{X}_k = [\mathbf{x}_{k1},\dots,\mathbf{x}_{kN}] \in \mathbb{R}^{d\times N}$. The associated distributions $\mu_{\mathbf{X}_k}$ are defined as previously.
A barycenter of these distributions in the Wasserstein metric is defined as
\begin{equation}
\mu_{\text{Bar}} = \underset \mu \argmin \sum_{k=1}^K w_k W_p(\mu,\mu_{\mathbf{X}_k})^p,
\label{wass_bar}
\end{equation}
for some positive weights $w_k$ verifying $\sum_{k=1}^K w_k = 1$. Somehow, this generalizes the concept of displacement interpolation for an arbitrary finite number of distributions. In various applications Wasserstein barycenters appear to be more suitable than Euclidean or more sophisticated barycenters (\cite{cuturi_1},\cite{gramf}); in particular they are robust to shifts and "elastic" deformations.
As previously, in 1D (i.e. $d=1$) and for $p=2$, this barycenter is known in closed-form. It is the uniform discrete probability measure defined over the set
\begin{equation}
\{\mathbf{x}_{\text{Bar}i} = \sum_{k=1}^K w_k \mathbf{x}_{k}[\sigma_k(i)],i=1\dots N\},
\label{1dwass_bar} 
\end{equation}
where $\sigma_k$ is a permutation in $\Sigma_N$ which sorts $\mathbf{x}_k$ entries (i.e. $\mathbf{x}_{k}[\sigma_k(1)]\leq \dots \leq \mathbf{x}_{k}[\sigma_k(N)]$).

For $d>1$, computing this barycenter is in general intractable\cite{gangbo}. Therefore in this paper we consider the Sliced Wasserstein Barycenter defined by replacing the Wasserstein distance by its sliced counterpart in the Wasserstein barycenter definition:
\begin{equation}
\mu_{\text{SBar}} = \underset \mu \argmin \sum_{k=1}^K w_k SW_p(\mu,\mu_{\mathbf{X}_k})^p.
\label{swass_bar}
\end{equation}

\textit{We use p = 2 in the following and in practice.}

\subsection{Data representation in the Transport Framework}
\label{data_rep}
We consider a $N_l \times N_c$ gray levels image.
We denote $\mathbf{x} = (x_i)_{1\leq i\leq N}$ the vector of pixel intensities rearranged in lines lexicographic order, with $N=N_lN_c$. The image is treated as a point cloud defined as 
\begin{equation}
\mathbf{X} = [\mathbf{v}_1,\cdots,\mathbf{v}_N],\text{with }\mathbf{v}_i = [x_i,\floor{\frac{i-1}{N_c}},(i-1(\text{mod}N_c))]^T.
\label{im_mod}
\end{equation}
This point cloud can be viewed as a representation of the considered image as a discrete surface in $\mathbb{R}^3$.
Given a set of unresolved objects images, the idea in the following section will be to match and or compute barycenters of the associated surfaces based on sliced transport. 

We use a ground metric $\mathcal{C}$ taking the following form: for two vectors $\mathbf{p} = [p_1,p_2,p_3]^T$ and $\mathbf{q} = [q_1,q_2,q_3]^T$ in $\mathbb{R}^3$, 
\begin{equation}
\mathcal{C}^2(\mathbf{p},\mathbf{q}) = (p_1-q_1)^2+\beta^2*((p_2-q_2)^2+(p_3-q_3)^2),
\label{gnd}
\end{equation}
where $\beta$ is a strictly positive real. As a reminder, in the applications, the first components $p_1$ and $q_1$ will be pixel intensities and the two other components will be pixel positions. For two images $\mathbf{x} = (x_i)_{1\leq i\leq N}$ and $\mathbf{y} = (y_i)_{1\leq i\leq N}$, let $\mathbf{X}$ and $\mathbf{Y}$ be the associated point clouds and $\sigma^*$ an optimal assignment of $\mathbf{Y}$'s points to $\mathbf{X}$'s points: then 
\begin{dmath}
W_2(\mu_{\mathbf{X}},\mu_{\mathbf{Y}})^2 = \sum_{i=1}^{N} (x_i-y_{\sigma^*(i)})^2 \\
+\beta^2*((\floor{\frac{i-1}{N_c}}-\floor{\frac{\sigma^*(i)-1}{N_c}})^2+((i-1(\text{mod}N_c))-(\sigma^*(i)-1(\text{mod}N_c)))^2). 
\label{wasser_dist_im}
\end{dmath}
Thus, if $\sigma^*$ is the identity, $W_2(\mu_{\mathbf{X}},\mu_{\mathbf{Y}}) = \|\mathbf{x}-\mathbf{y}\|_2$. Therefore, we can consider $W_2$ (and $SW_2$) as a generalization of the euclidian distance in the pixels domain. 

Conversely, from a given point cloud $\mathbf{Z} \in \mathbb{R}^{3\times N}$, we can go backward to an image by discretizing the 2D function $\Phi[\mathbf{Z}]$ defined as 
\begin{equation}
\Phi[\mathbf{Z}](x,y) = \sum_{i=1}^N \mathbf{Z}[1,i]\delta_{\mathbf{Z}[2,i],\mathbf{Z}[3,i]}(x,y).
\label{im_point_cloud}
\end{equation}    
\subsection{Alternative image registration frameworks}
Numerical optimal transport can be casted as a non-rigid registration tool and has been used for this purpose in medical imaging for instance (see \cite{haker2004optimal}). It is somehow close to optical flow estimation based registration methods; a comprehensive review of which can be found in \cite{fortun2015optical}. Indeed in the latter framework, the registration relies on the estimation of a motion field which transforms a reference image into a target image. However optical flow estimation methods classically assume pixels brightness constancy as they move or "high order" constancy to account for illumination changes between the reference and the target images.
These hypothese are relevant whenever the warping can be described as resulting from locally rigid shifts i.e. when there are solid objects moving in a scene.  This is not true in the present application since the PSFs evolution across the FOV consists in a complex warping in intensity and space directions.

However authors have proposed to replace the brightness constancy constraint with a "mass" preservation constraint which is better suited for fluids-like motions (see \cite{OEOF, fire_detect, mueller2013optical, dense_flow_est}).  
In this framework, the optical flow estimation problem turns into an optimal transport problem if the chosen regularization of the motion field consists in minimizing the associated kinetic energy. 

\section{Transport based PSF field interpolation}
\label{train_meth}
We consider a set of $K$ PSFs $\mathcal{S} = (\mathbf{x}_k)_{1\leq k\leq K}$, located at the positions $(\mathbf{u}_k)_{1\leq k\leq K}$ in the fov. We note $(\mathbf{X}_k)_{1\leq k\leq K}$ the associated point clouds as defined in Section \ref{data_rep}. We want to estimate the PSF at a new location $\mathbf{u}$. We note this PSF $\mathbf{x}_{\mathbf{u}}$ and we note $\mathbf{X}_{\mathbf{u}}$ the associated point cloud:
\begin{equation}
\mathbf{x}_{\mathbf{u}} \equiv \Phi[\mathbf{X}_{\mathbf{u}}],
\label{im_mapping}
\end{equation}
where the operator $\Phi$ is defined in Eq.~\ref{im_point_cloud}. 
We want to estimate $\mathbf{X}_{\mathbf{u}}$ as a Wassertein barycenter of a subset of PSFs indexed by $\mathcal{I}(\mathbf{u})$ in the neighborhood of $\mathbf{u}$ in the fov:
\begin{equation}
\mathbf{X}_{\mathbf{u}} = \underset{\mathbf{X}}\argmin \sum_{k \in \mathcal{I}(\mathbf{u})} w_k(\mathbf{u}) SW_2(\mu_{\mathbf{X}},\mu_{\mathbf{X}_k})^2,
\label{interp_mod} 
\end{equation}
with $w_k(\mathbf{u})\geq 0$ and $\sum_{k\in \mathcal{I}(\mathbf{u})} w_k(\mathbf{u})=1$. 
$\mathcal{I}(\mathbf{u})$ simply indexes the nearest neighbors of $\mathbf{u}$ in the set of locations $(\mathbf{u}_k)_{1\leq k\leq K}$. The question of the size of this neighborhood is postponed to Section \ref{param}.

The weights $(w_k(\mathbf{u}))_{k\in \mathcal{I}(\mathbf{u})}$ can be seen as generalized barycentric coordinates of $\mathbf{X}_{\mathbf{u}}$ relatively to the clouds $(\mathbf{X}_k)_{k\in \mathcal{I}(\mathbf{u})}$, in the Wassertein metric.  

These weights are calculated in three steps that we detail in the following sections:
\begin{itemize}
\item \underline{Embedding}: we calculate the pairwise approximated Wasserstein 2 distances over the set $(\mathbf{X}_k)_{k\in \mathcal{I}(\mathbf{u})}$ and determine an Euclidean embedding of $(\mathbf{X}_k)_{k\in \mathcal{I}(\mathbf{u})}$ that preserves the Wasserstein distances; we get a set of coordinates $(\mathbf{r}_k)_{1\leq k \leq |\mathcal{I}(\mathbf{u})|}$, where $|\mathcal{I}(\mathbf{u})|$ is the number of elements in $\mathcal{I}(\mathbf{u})$;
\item \underline{Interpolation}: we estimate the representation $\mathbf{r}_{\mathbf{u}}$ of $\mathbf{X}_{\mathbf{u}}$ in the previously calculated embedding by interpolating the set $(\mathbf{r}_k)_{1\leq k \leq |\mathcal{I}(\mathbf{u})|}$ coordinates-wise;
\item \underline{Weights setting}: the weights $(w_k(\mathbf{u}))_{k\in \mathcal{I}(\mathbf{u})}$ are calculated as the Euclidean barycentric coordinates of $\mathbf{r}_{\mathbf{u}}$ relatively to $(\mathbf{r}_k)_{1\leq k \leq |\mathcal{I}(\mathbf{u})|}$.
\end{itemize}

\subsection{Local non-linear dimension reduction}
\label{embedding}
We recall that we want to estimate the PSF at a position $\mathbf{u}$ in the fov, given the PSFs located at the positions $(\mathbf{u}_k)_{1\leq k\leq K}$. We consider the $p$ nearest neighbors of $\mathbf{u}$ in $(\mathbf{u}_k)_{1\leq k\leq K}$. This defines the set $\mathcal{I}(\mathbf{u})$ aforementioned.
 
As we will see in the Numerical experiments Section, the parameter $\beta$ in Eq.~\ref{wasser_dist_im} is set so that for the closest PSFs in the fov, the Wassertein distance is equal to the euclidian distance. Indeed, the Euclidean distance is usually assumed to be a good approximation of the geodesic distance between close points on a given Manifold (see for example \cite{tenenbaum_global_2000,Donoho03hessianeigenmaps}). However, the more two PSFs are distant in the fov, the more the PSFs manifold curvature manifests through the relative warping of their structures. By minimizing the amount of work needed to push one of the PSFs toward the other, the Wasserstein metric can keep track of the warping to a certain extent, thus unfolding locally the Manifold. For this reason, the Wasserstein distance can potentially give a faithful approximation of the geodesic distances, on broader neighborhoods than the Euclidean distance. 
We define the $p\times p$ local pairwise distance matrix $\mathbf{M}_W(\mathbf{u})$ as:
\begin{equation}
\mathbf{M}_W(\mathbf{u})[i,j] \approx W_2(\mu_{\mathbf{X}_{\mathcal{I}(\mathbf{u})[i]}},\mu_{\mathbf{X}_{\mathcal{I}(\mathbf{u})[j]}})^2, \; (i,j) \in \llbracket 1,p \rrbracket^2.
\label{local_dist_mat}
\end{equation}

Then we apply the Multidimensional Scaling (MDS) procedure to calculate the embedding\cite{Abdi2007}. The first step consists in converting the distance matrix $\mathbf{M}_W(\mathbf{u})$ into a Gram matrix  $\mathbf{X}_i$ i.e. whose entries are given by $\mathbf{X}_i[i,j] = \langle \mathbf{r}_i,\mathbf{r}_j \rangle$, so that $\mathbf{M}_W(\mathbf{u})[i,j] = \|\mathbf{r}_i-\mathbf{r}_j\|_2$. The vector $\mathbf{r}_i$ is the low dimensional embedding of $\mathbf{X}_{\mathcal{I}(\mathbf{u})[i]}$.

For this purpose we introduce the "centering matrix" \cite{mds} defined as 
\begin{equation}
\mathbf{C} = \mathbf{I}_p - \frac{1}{p}\mathds{1}_p\mathds{1}_p^T,
\label{cent_mat}
\end{equation}
where $\mathds{1}_p$ is a row vector made of $p$ ones. 
Multiplying a matrix by $\mathbf{C}$ on the left has the effect of subtracting to each column its mean; $\mathbf{C}$ being symmetric, multiplying a matrix by $\mathbf{C}$ on the right subtracts to each line its mean. 

The Gram matrix is calculated as
\begin{equation}
\mathbf{X}_i = - \frac{1}{2} \mathbf{C}\mathbf{M}_W(\mathbf{u})\mathbf{C}, 
\label{cross_prod_cv}
\end{equation}
where $\mathbf{M}_W(\mathbf{u})$ is squared entry-wise. 
We define the embedding coordinates matrix $\mathbf{R} = [\mathbf{r}_1,\dots,\mathbf{r}_n]$. Assuming that such an embedding exists and $\mathbf{R}$ has null lines and columns means, one can show that $\mathbf{X}_i = \mathbf{R}^T\mathbf{R}$ (see appendix \ref{mds_cent}).
Under this hypothesis, $\mathbf{X}_i$ is a symmetric matrix and therefore can be orthogonally diagonalized:
\begin{equation}
\mathbf{X}_i = \mathbf{V}^T\mathbf{S}\mathbf{V},
\label{g_svd}
\end{equation}
where $\mathbf{V}^T\mathbf{V} = \mathbf{V}\mathbf{V}^T = \mathbf{I}_p$ and $\mathbf{S}$ is a non-negative diagonal matrix. 
It follows that $\mathbf{R}$ can be calculated as 
\begin{equation}
\mathbf{R} = \mathbf{Q}\mathbf{S}^{\frac{1}{2}}\mathbf{V},
\label{embedding_calculation}
\end{equation}
where $\mathbf{Q}$ can be any orthogonal matrix; we set it to the identity in practice \cite{dokmanic2015Euclidean}. 
Depending on neighborhood's size $p$, the last diagonal values of $\mathbf{S}$ can be neglected so that the last lines of $\mathbf{V}$ can be discarded; besides this represents a way of analyzing locally the Manifold dimensionality \cite{tenenbaum_global_2000}. 

We note d the dimensionality of the calculated embedding in Algorithm \ref{TraIn}.

\subsection{Field-of-view mapping}
\label{fov_map}
In this section, we estimate the local low dimensional embedding of the unknown PSF located at $\mathbf{u}$ in the fov. We note this embedding $\mathbf{r}_{\mathbf{u}}$. We note $d$ the dimension of the vectors $(\mathbf{r}_i)_{1\leq i\leq p}$ determined in the previous section, and $\mathbf{r}_{\mathbf{u}}$. To compute the $i^{th}$ component of $\mathbf{r}_{\mathbf{u}}$ we determine an interpolating function 
\begin{equation}
f_i: \mathbb{R}^2 \rightarrow \mathbb{R}\; / f_i(\mathbf{u}_j) = \mathbf{r}_j[i], \; \forall j \in \llbracket 1,p\rrbracket.
\label{exact_interp} 
\end{equation}
This is a standard surface interpolation problem that we solve using the so-called thin-plate spline\cite{eberly2002thin} which is appealing because of its physical interpretation: it is the exact interpolating function that minimizes the "bending energy" defined as:
\begin{equation}
E(f) = \int_{\mathbb{R}^2} (\frac{\partial^2 f(x,y)}{\partial^2 x})^2+2*(\frac{\partial^2 f(x,y)}{\partial x\partial y})^2+(\frac{\partial^2 f(x,y)}{\partial^2 y})^2\text{d}x\text{d}y. 
\label{bend_en}
\end{equation} 

Thus, the function $f_i$ takes the following form:
\begin{equation}
f_i(\mathbf{x}) = a x + b y + c + \sum_{j=1}^p a_{ij} \|\mathbf{x}-\mathbf{u}_j\|_2^2 \text{ln}(\|\mathbf{x}-\mathbf{u}_j\|_2),
\label{thin_plate_rbf} 
\end{equation}
with $\mathbf{x} = (x,y)$ and the coefficients a, b, c and $a_{ij}$ are calculated so that $f_i$ takes the prescribed values at the control points $(\mathbf{u}_j)_{1\leq j\leq p}$.
This way, each component of $\mathbf{r}_{\mathbf{u}}$ can be estimated. 
 
\subsection{Barycentric coordinates}
The embedding of the PSF at the position $\mathbf{u}$ in the fov has been determined. Now, we want to estimate the point cloud $\mathbf{X}_{\mathbf{u}}$ from this embedding and the neighbor PSFs. In other terms, we need to determine the weights $w_k(\mathbf{u})$ in Eq.~\ref{interp_mod}. 
In Section \ref{embedding}, we calculated an isometric embedding of the matrices $(\mathbf{X}_k)_{k\in \mathcal{I}(\mathbf{u})}$. Because of this isometry, for a set of positive weights $(w_i)_{1\leq i\leq p}$ verifying $\sum_{i=1}^p w_i = 1$, the problem 
\begin{equation}
\underset{\mathbf{x}}{{\text{min}}} \sum_{i=1}^p w_i\|\mathbf{x}-\mathbf{r}_i\|_2^2
\label{bar_emb}
\end{equation}
is equivalent to the following 
\begin{equation}
\underset{\mathbf{X}}{{\text{min}}} \sum_{i=1}^p w_i SW_2(\mu_{\mathbf{X}},\mu_{\mathbf{X}_{\mathcal{I}(\mathbf{u})[i]}})^2.
\label{bar_sw}
\end{equation}

This provides a mean for computing $\mathbf{X}_{\mathbf{u}}$ from $\mathbf{r}_{\mathbf{u}}$.
Indeed, we first can consider the following barycentric coordinates problem:
\begin{equation}
\underset{w_1,\dots,w_p} {{\text{min}}} \frac{1}{2}\|\mathbf{r}_{\mathbf{u}} - \sum_{i=1}^p w_i\mathbf{r}_i\|_2^2 \; \text{s.t.} \; w_i\geq 0 \text{ and } \sum_{i=1}^p w_i = 1.
\label{bar_coord_pb} 
\end{equation}
We note $(w_i^*)_{1\leq i\leq p}$ the optimal tuple and we define $\mathbf{r}_{\mathbf{u}}^* = \sum_{i=1}^p w_i^*\mathbf{r}_i$.  
$ \mathbf{r}_{\mathbf{u}} \approx  \mathbf{r}_{\mathbf{u}}^*$ and $\mathbf{r}_{\mathbf{u}}^*$ is solution of the Problem \ref{bar_emb}, with the weights $(w_i^*)_{1\leq i\leq p}$. Therefore, we compute $\mathbf{X}_{\mathbf{u}}$ as
\begin{equation}
\mathbf{X}_{\mathbf{u}} = \underset{\mathbf{X}}\argmin \sum_{i=1}^p w_i^* SW_2(\mu_{\mathbf{X}},\mu_{\mathbf{X}_{\mathcal{I}(\mathbf{u})[i]}})^2.
\label{backward_trans}
\end{equation}
This approach is in a way similar to the procedure described in \cite{gower}.
The final image is obtained by discretizing the function $\Phi[\mathbf{X}_{\mathbf{u}}]$ defined by Eq.~\ref{im_point_cloud}. This step is discussed in Section \ref{practical}. 

\subsection{Algorithm}
\label{algo}
\begin{figure}
\begin{center}
\includegraphics[scale=0.3]{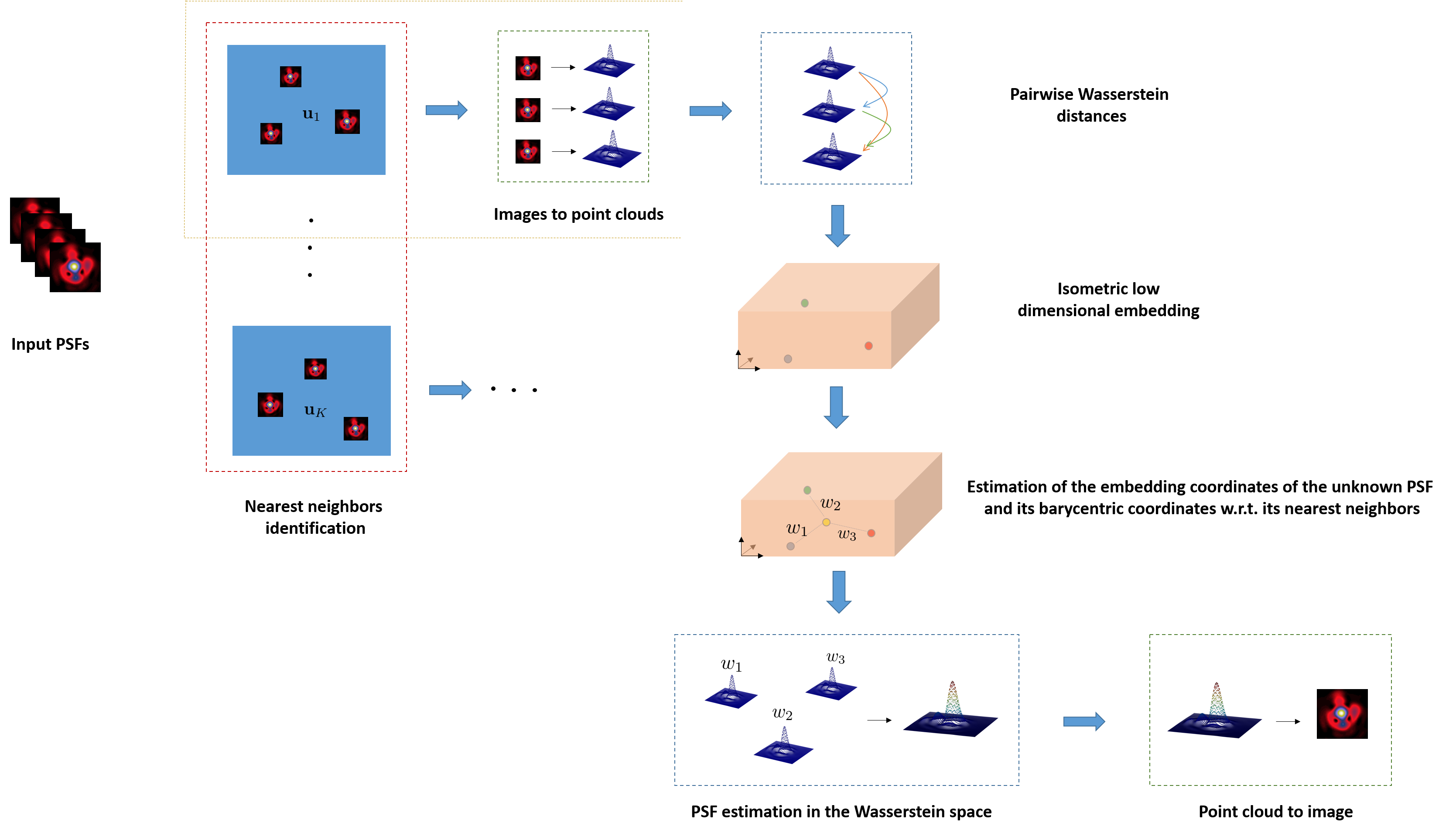}
\caption{TraIn procedure illustration.}
\end{center}
\label{train_im}
\end{figure}

We want to estimate $D$ PSFs at random locations $(\mathbf{v}_i)_{1\leq i\leq D}$ from a set of $K$ PSFs at known locations in the fov (see Section \ref{train_meth}). The whole procedure is summarized in Algorithm $\ref{TraIn}$ and illustrated in Fig.~\ref{train_im}.

\begin{algorithm*}[!htb]
\caption{Transport Interpolation (TraIn)}
\begin{algorithmic}[1]
\label{TraIn}

\bigskip
\STATE{\bf Inputs:} $K$ PSFs $(\mathbf{x}_k)_{1\leq k \leq K}$, $K$ observations locations $(\mathbf{u}_k)_{1\leq k \leq K}$, $D$ interpolation locations $(\mathbf{v}_k)_{1\leq k \leq D}$, number of neighbors $p$, local dimensionality $d\leq p$
\STATE Compute the weighting parameter $\beta$ (see Eq.~\ref{wasser_dist_im}) 
\STATE Compute the $p$ neighbors of each location $\mathbf{v}_i$ in the set $(\mathbf{u}_k)_{1\leq k \leq K}$; this results in a collection of sets of indices $(\mathcal{I}(\mathbf{v}_i))_{1\leq i \leq D}$ 
\STATE Transform the PSFs into point clouds (see Eq.~\ref{im_mod}): $(\mathbf{x}_k)_{1\leq k \leq K} \shortrightarrow (\mathbf{X}_k)_{1\leq k \leq K}$  
\STATE Compute the approximated Wassertein 2 distances between pairs of point clouds $(\mathbf{X}_l, \mathbf{X}_m)\; / (l,m) \in \mathcal{I}(\mathbf{v}_i)^2$ for some $i \in \llbracket 1,D\rrbracket$  
 \FOR{$i=1$ to $D$}
 	\STATE Form the pairwise Wasserstein distances matrix over the set $(\mathbf{X}_j)_{j\in \mathcal{I}(\mathbf{v}_i)}$ and compute a local euclidian embedding; the result is a set of vectors $(\mathbf{r}_j)_{1\leq j\leq p}$ in $\mathbb{R}^d$ 
 	\STATE Estimate the embedded coordinates at the location $\mathbf{v}_i$ using a thin-plate spline interpolation coordinate-wise ; this results in a vector $\mathbf{r}_{\mathbf{v}_i}$
 	\STATE Compute the barycentric coordinates of $\mathbf{r}_{\mathbf{v}_i}$ relatively to the vectors $(\mathbf{r}_j)_{1\leq j\leq p}$ 
 	\STATE Compute the approximate Wassertein barycenter of $(\mathbf{X}_j)_{j\in \mathcal{I}(\mathbf{v}_i)}$ using the previously calculated barycentric coordinates as weights 
 	\STATE Compute the interpolated PSF $\mathbf{x}_{\mathbf{v}_i}$ from this barycenter (see Eq.~\ref{im_point_cloud}) 
 \ENDFOR \\
 
 \STATE {\bf Return:} $(\mathbf{x}_{\mathbf{v}_i})_{1\leq i\leq D}$.
\end{algorithmic}
\end{algorithm*}

This procedure is generic in the sense that the Wassertein metric can be replaced by an arbitrary metric provided that one is able to compute geodesic distances and geodesics as we will see in Section \ref{practical}.

\section{Numerical results}
\label{num_res}
\begin{figure}
\begin{center}
\includegraphics[scale=0.35]{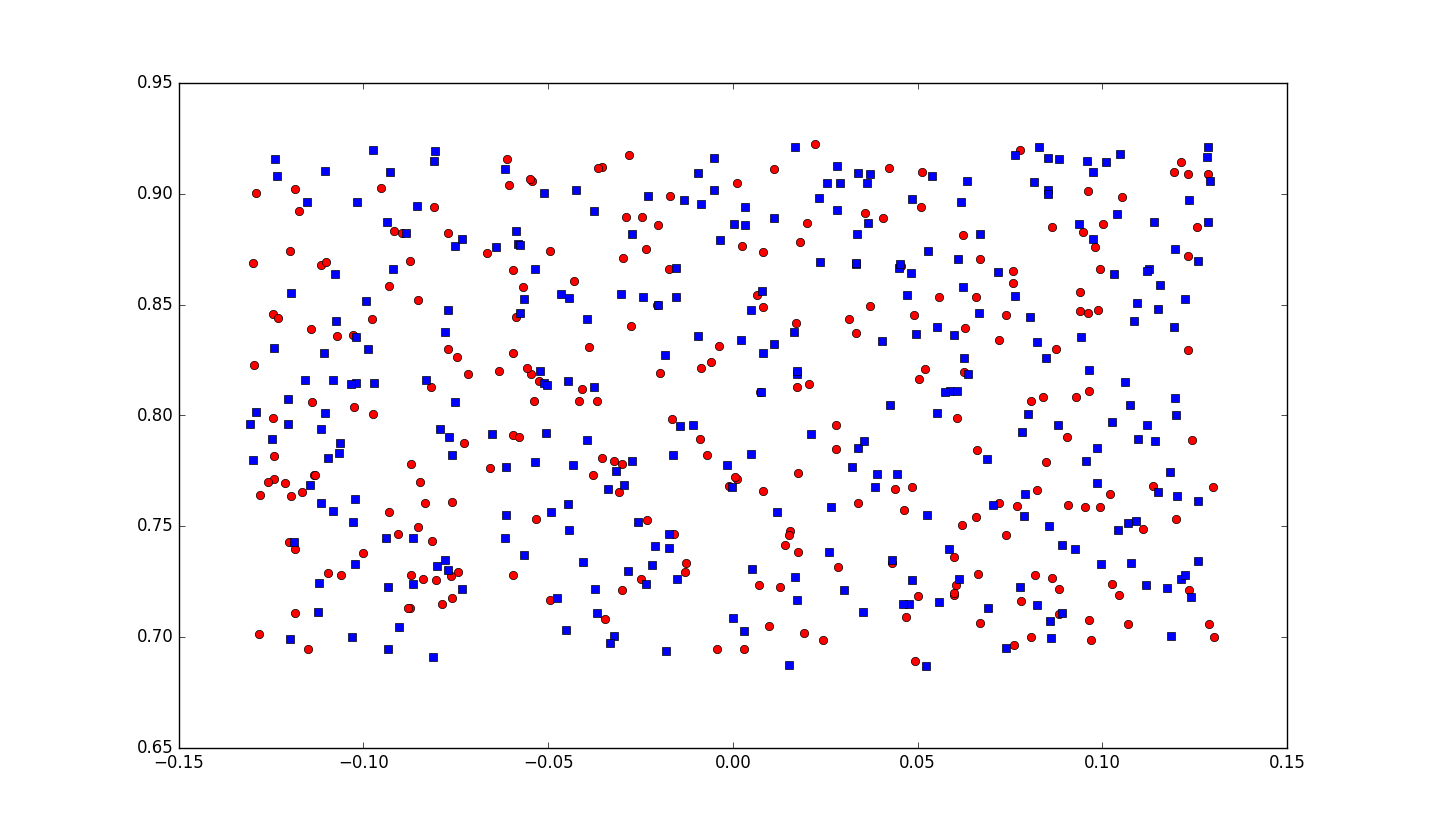}
\end{center}
\caption{Simulated PSFs distribution across the FOV; the blue squares represent the 300 observed PSFs central locations and the red circles the 250 PSFs to be interpolated; the coordinates are in degrees.}
\label{distrib}
\end{figure}

We tested the proposed method on a set of 550 simulated Euclid telescope optical PSFs as in \cite{fng}. 
The PSFs are distributed in the fov according to Fig.~\ref{distrib}. We split the data into a "learning set" made of 300 observed PSFs and a "test set" corresponding to 250 unknown PSFs. 

\subsection{Quality assessment}
Astronomical imaging applications such as weak lensing and galaxies classification rely on characterizing galaxies shape and luminosity\cite{Crop1}. Therefore the PSF's shape is crucial. 
We recall the definition of the central moments of an image $\mathbf{X} = (x_{ij})_{i,j}$: 
\begin{equation}
\mu_{s,t}(\mathbf{X}) = \sum_i\sum_j(i-i_c)^s(j-j_c)^t x_{ij} 
\end{equation}
with $(s,t) \in \mathbb{N}^2$, $(i_c,j_c)$ being the image centroid coordinates.

The PSF's shape can be characterized by the so-called "ellipticity parameters"\cite{kaiser} defined as follows:
\begin{gather}
e_1(\mathbf{X}) = \frac{\mu_{2,0}(\mathbf{X})-\mu_{0,2}(\mathbf{X})}{\mu_{2,0}(\mathbf{X})+\mu_{0,2}(\mathbf{X})}  \\
e_2(\mathbf{X}) = \frac{2\mu_{1,1}(\mathbf{X})}{\mu_{2,0}(\mathbf{X})+\mu_{0,2}(\mathbf{X})}.
\label{ellipticity}
\end{gather}

The vector $\boldsymbol{\gamma}(\mathbf{X}) = [e_1(\mathbf{X}),e_2(\mathbf{X})]^T$ tells how much $\mathbf{X}$ departs from an isotropic distribution of luminosity and gives one the main direction of elongation. 

Additionally, the PSF's morphology can be characterized by the average intensity-weighted distance of the pixels to the image centroid. We refer to this quantity as the PSF's size and it is calculated as follows:
\begin{equation}
\text{S}(\mathbf{X}) = (\frac{\sum_i\sum_j((i-i_c)^2+(j-j_c)^2) x_{ij}}{\sum_i\sum_j x_{ij}})^{1/2}. 
\end{equation}

We note $(\mathbf{\text{Im}}_i)_{1\leq i\leq D}$ the set of test PSFs and $(\widehat{\mathbf{\text{Im}}}_i)_{1\leq i\leq D}$ the set of corresponding PSFs interpolated with a given method. The reconstruction quality is accessed through the following quantities:
\begin{itemize}
\item the average error on the ellipticity vector: $\text{E}_{\boldsymbol{\gamma}} = \sum_{i=1}^D \|\boldsymbol{\gamma}(\mathbf{\text{Im}}_i) - \boldsymbol{\gamma}(\widehat{\mathbf{\text{Im}}}_i)\|_2/D$;
\item the average absolute error on the size: $\text{E}_{\text{S}} = \sum_{i=1}^D |\text{S}(\mathbf{\text{Im}}_i) - \text{S}(\widehat{\mathbf{\text{Im}}}_i)|/D$ in pixels;
\item the average normalized mean square error: $\text{NMSE} = \sum_{i=1}^D \|\widehat{\mathbf{\text{Im}}}_i-\mathbf{\text{Im}}_i\|_2^2/(D\|\mathbf{\text{Im}}_i\|_2^2)$
\end{itemize}

\subsection{Experiments}
We compare the proposed method called TraIn to the Inverse Distance Weighting (IDW) and Radial Basis Function (RBF) based interpolation. These methods are described in \cite{gentile_2013} as the ones which perform the best on the considered PSFs interpolation. For the first method, the weights are calculated based on squared distances in the FOV; this is often referred as "Inverse distance-squared interpolator". We used the radial basis function $f(r) = r^2\text{ln}(r)$; thus the second method is nothing but a thin plate spline interpolation as described in Section \ref{fov_map}. The $300$ observed PSFs are first decomposed using a principal components analysis and retaining 40 principal components. This yields a representation with 40 coefficients for each PSF. These coefficients are interpolated component-wise with the two comparison methods and the interpolated PSFs are derived. As the method proposed, the IDW and RBF based interpolation methods are local in the sense that a PSF is interpolated at a given location in the FOV based on a given number of PSFs observed in the vicinity of the considered location. Thus, the three methods are applied using different numbers of "neighbors PSFs". 

\begin{figure}
\begin{center}
\includegraphics[scale=0.45]{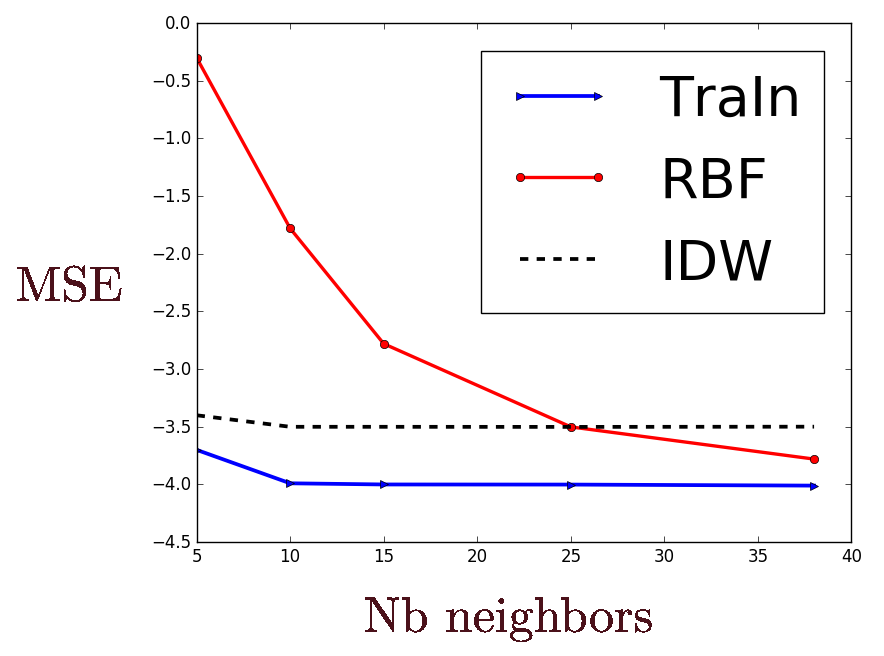}
\end{center}
\caption{Normalized mean squared error: Y-axis $\text{log}_{10}(\text{NMSE})$, X-axis number of neighbors.}
\label{mse}
\end{figure} 

\begin{figure}
\begin{center}
\includegraphics[scale=0.45]{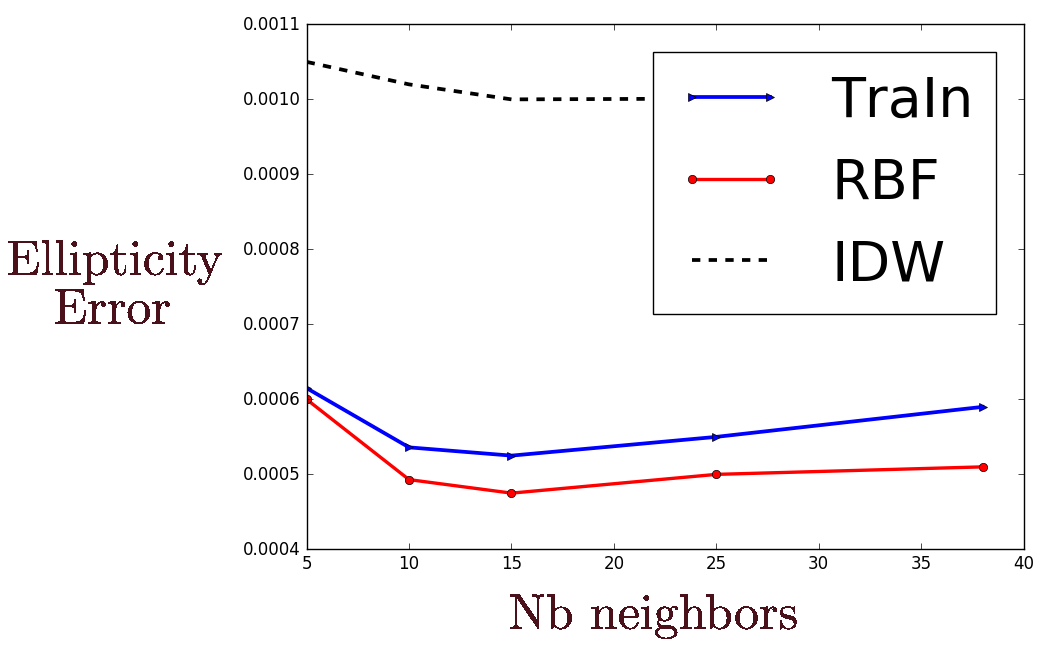}
\end{center}
\caption{Average error on the ellipticity: Y-axis $\text{E}_{\boldsymbol{\gamma}}$, X-axis number of neighbors.}
\label{err_ell}
\end{figure} 

\begin{figure}
\begin{center}
\includegraphics[scale=0.45]{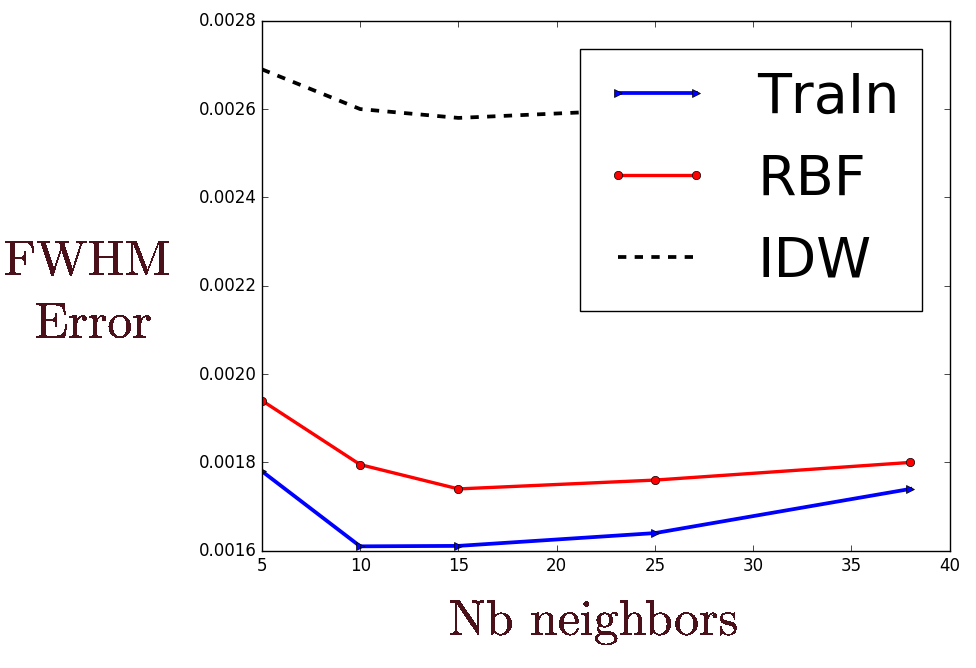}
\end{center}
\caption{Average absolute error on the size: Y-axis $\text{E}_{\text{S}}$, X-axis number of neighbors.}
\label{err_fwhm}
\end{figure}
\begin{figure}
\begin{center}
\includegraphics[scale=0.45]{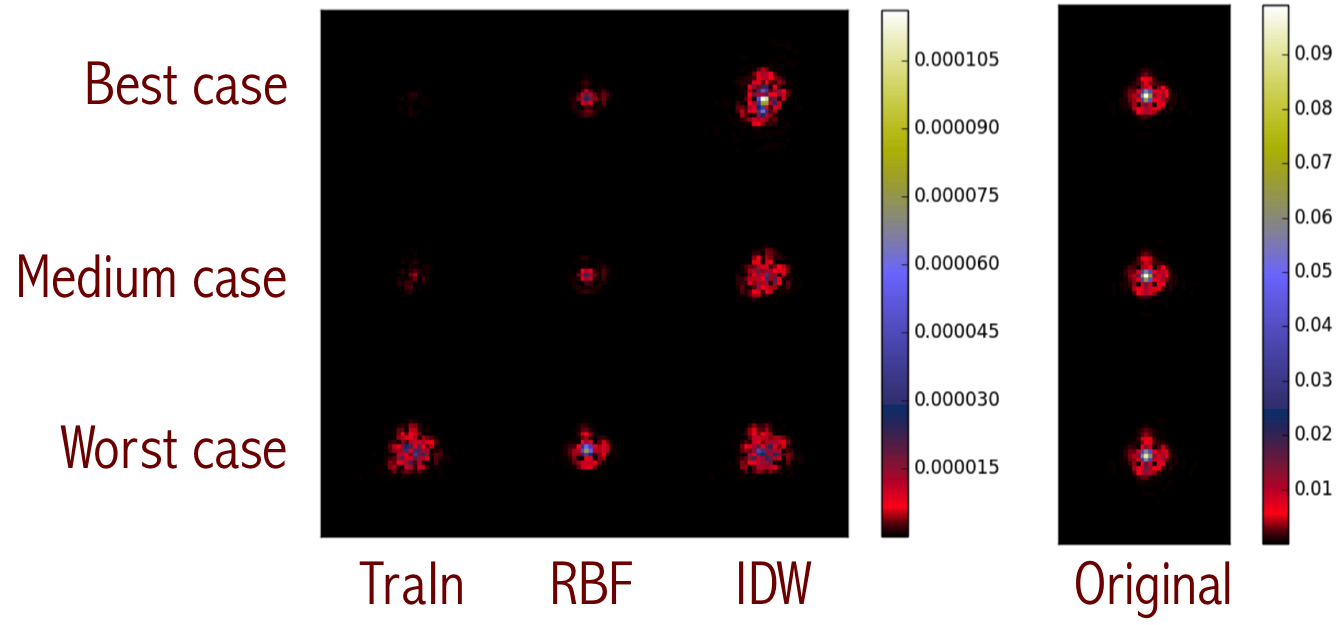}
\end{center}
\caption{Absolute error images for 15 neighbors; from the left to the right, TraIn, RBF, IDW, original image; from the top to the bottom, best, average and worst case in terms of the accuracy of TraIn with respect to the pixels mean square error.}
\label{res_examples}
\end{figure}

 
The results are shown in Fig.~\ref{mse},\ref{err_ell},\ref{err_fwhm}. Interestingly, the way the methods compare is quite different depending on the criterium considered. This will be discussed in the next section. However, the less accurate method appears to be the IDW based method. Indeed, it gives the less accurate results in terms of shape; in particular, it is almost an order of magnitude less accurate than the two other methods for the ellipticity. Furthermore, it is almost constantly less accurate than TraIn with respect to the pixels mean square error. The RBF based method and the proposed one gives comparable errors with respect to the shape parameters; it is slightly more accurate on the ellipticity and slightly less accurate on the shape. The lowest errors with the RBF based method are obtained for a number neighbors around 15 for the shape parameters. The proposed method has a quite stable accuracy with respect to the number of neighbors for all the criteria. Moreover, it is typically several orders of magnitude more accurate than the RBF based one on the pixels mean square errors for a number of neighbors smaller than 15 (see Fig.~\ref{mse}). Therefore, the TraIn method is globally the most accurate. 
We give some examples of absolute error images i.e. $|\mathbf{\text{Im}}_i - \widehat{\mathbf{\text{Im}}}_i|$ for some $i \in \llbracket 1,N \rrbracket$ in Fig.~\ref{res_examples}, which correspond to interpolations with 15 neighbors. It shows indeed that TraIn typically yields a substantially lower residual. 

\subsection{Discussion}
\begin{figure}
\begin{center}
\includegraphics[scale=0.55]{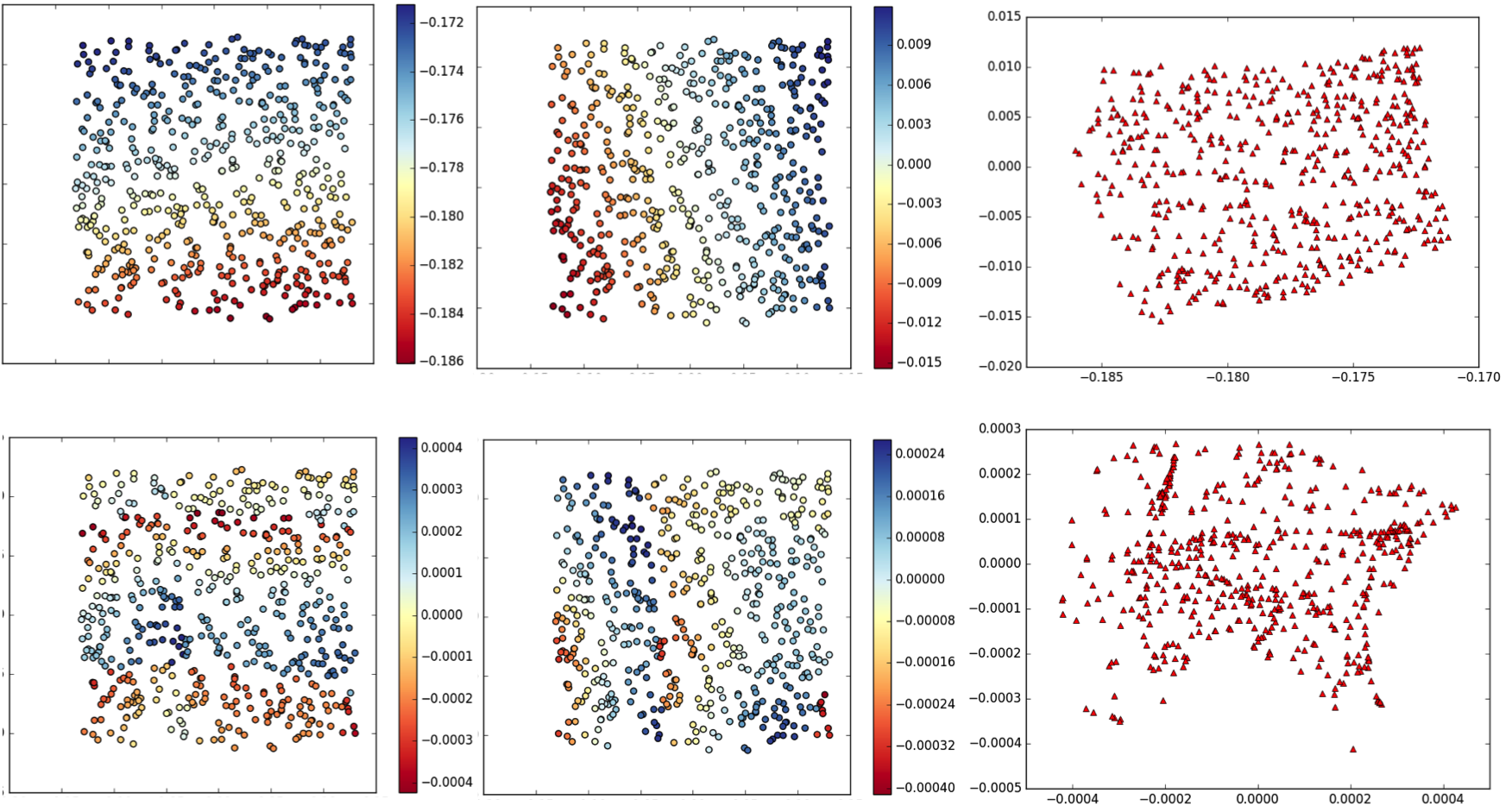}
\end{center}
\caption{Spatial distributions of the coefficients of the PSFs used for the experiments relative to 4 principal components calculated over the "learning" PSFs set; the top panel corresponds to the first and second principal components and the coefficients scatter plot;  the bottom panel corresponds to the 6th and 7th principal components and the coefficients scatter plot as well.}
\label{pca_coeff_distrib}
\end{figure}

In this Section, we discuss the performances of the tested methods. We consider the q first principal components derived from the learning PSFs set. In the numerical experiments, we set $q=40$ for the RBF based methods. The Figure \ref{pca_coeff_distrib} illustrates how coefficients magnitudes vary across the field for the whole data set (including the test PSFs) for some of the principal components. 
The two first components exhibit a smooth spatial evolution so that it is possible to interpolate accurately the corresponding surfaces with RBFs using few observations or control points in the vicinity of the interpolation point. This smoothness can be seen in the scatter plot associated with these two first components. Indeed, one needs a smooth warping to transform the scatter plot distribution into the PSFs spatial distribution. 
On the contrary, the magnitudes have sharper variations for the two other components. This can also be seen from the scatter plot, since a highly non-linear transform would be needed for mapping the scatter plot distribution into the PSFs spatial distribution. In this case, more control points are required for the RBF interpolation to be robust to fast local variations. 
The scatter plots of Fig.~\ref{pca_coeff_distrib} are nothing but orthogonal projections of the PSFs manifold over the vector plans spanned by the respective pairs of principal components. As the indexes of the principal components increase, the manifold complexity becomes more apparent and manifests in fast spatial variations of the representation coefficients. Thus, when a few control points are used, the RBF interpolation accumulates errors on the "high indexes" components resulting in a poor accuracy in terms of pixels values. TraIn and IDW instead interpolate globally the PSFs in the sense that the interpolation is not component-wise; this implies less dependency of pixel-wise accuracy on the number of control points and in particular a better accuracy compared to the RBF interpolation where a few control points are used. 

\begin{figure}
\begin{center}
\includegraphics[scale=0.45]{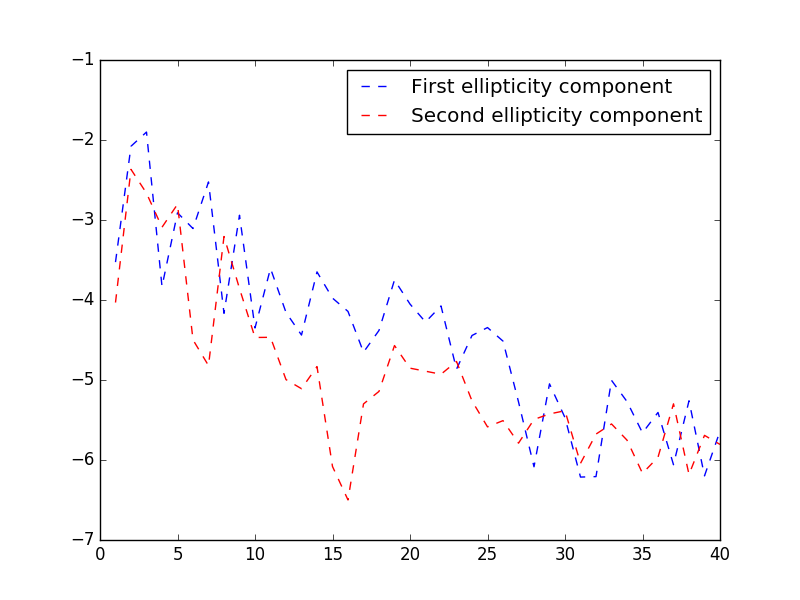}
\end{center}
\caption{Ellipticity components sensitivity to the PCA atoms. Y axis: $\text{log}_{10}(\mathcal{V}(e_k,\mathbf{P}_j))$(see Eq.~\ref{sens}); X axis: PCA atoms indexes $j$; moderate errors on the PCA coefficients have a weak impact on the ellipticity, especially for the "high indexes" PCA components.}
\label{ellip_sensitivity}
\end{figure}

However, this does not hold true for the ellipticity. To understand these changes of performances, one needs to consider the shape parameters sensitivity relatively to the principal components. Specifically, for each PSF $\mathbf{X}_i$ treated as an $N_l \times N_c$ matrix, we calculate the derivative of each ellipticity parameter along the lines passing though $\mathbf{X}_i$ and parallel to each of the principal components $\mathbf{P}_j$ also treated as matrices: $\frac{de_k(\mathbf{X}_i+t\mathbf{P}_j)}{dt}$, $k=1,2$. The analytic expressions can be found in Appendix \ref{ell_der}. For each principal component $\mathbf{P}_j$, we note $\text{disp}_j$ the dispersion of the PSFs set projected on the line passing through $\mathbf{0}$ and directed by $\mathbf{P}_j$. We define the sensitivity of the ellipticity component $e_k$, $k=1,2$ with respect to the principal component $\mathbf{P}_j$ as follows:
\begin{equation}
\mathcal{V}(e_k,\mathbf{P}_j) = \frac{\text{disp}_j}{(D+K)} \sum_i |\frac{d e_k(\mathbf{X}_i+t\mathbf{P}_j)(0)}{dt}|. 
\label{sens}
\end{equation} 
It measures how much small perturbations along each of the principal component is susceptible to make the ellipticity vary. In other words, it measures the sensitivity of the ellipticity to errors on the principal components coefficients. The sensitivities for the 40 first principal components are plotted in Fig.~\ref{ellip_sensitivity} for the two ellipticity components. It is interesting to see that the plots are not monotonically decreasing. We can draw an important remark from this observation: the most important features in terms of pixel-wise error are not necessarily the most influential in terms of shape. The ellipticity parameters have average magnitudes of order $10^{-2}$. Therefore, one can tell from Fig.~\ref{ellip_sensitivity} that the ellipticity is significantly sensitive to only a few principal components within the 10 first one. 
Therefore, as long as the coefficient-wise interpolation is accurate for those low indexes principal components while staying in a reasonable range in general, the final result is accurate in terms of ellipticity. This explains why the RBF interpolation maintains a good accuracy on the ellipticity.    

\begin{figure}
\begin{center}
\includegraphics[scale=0.35]{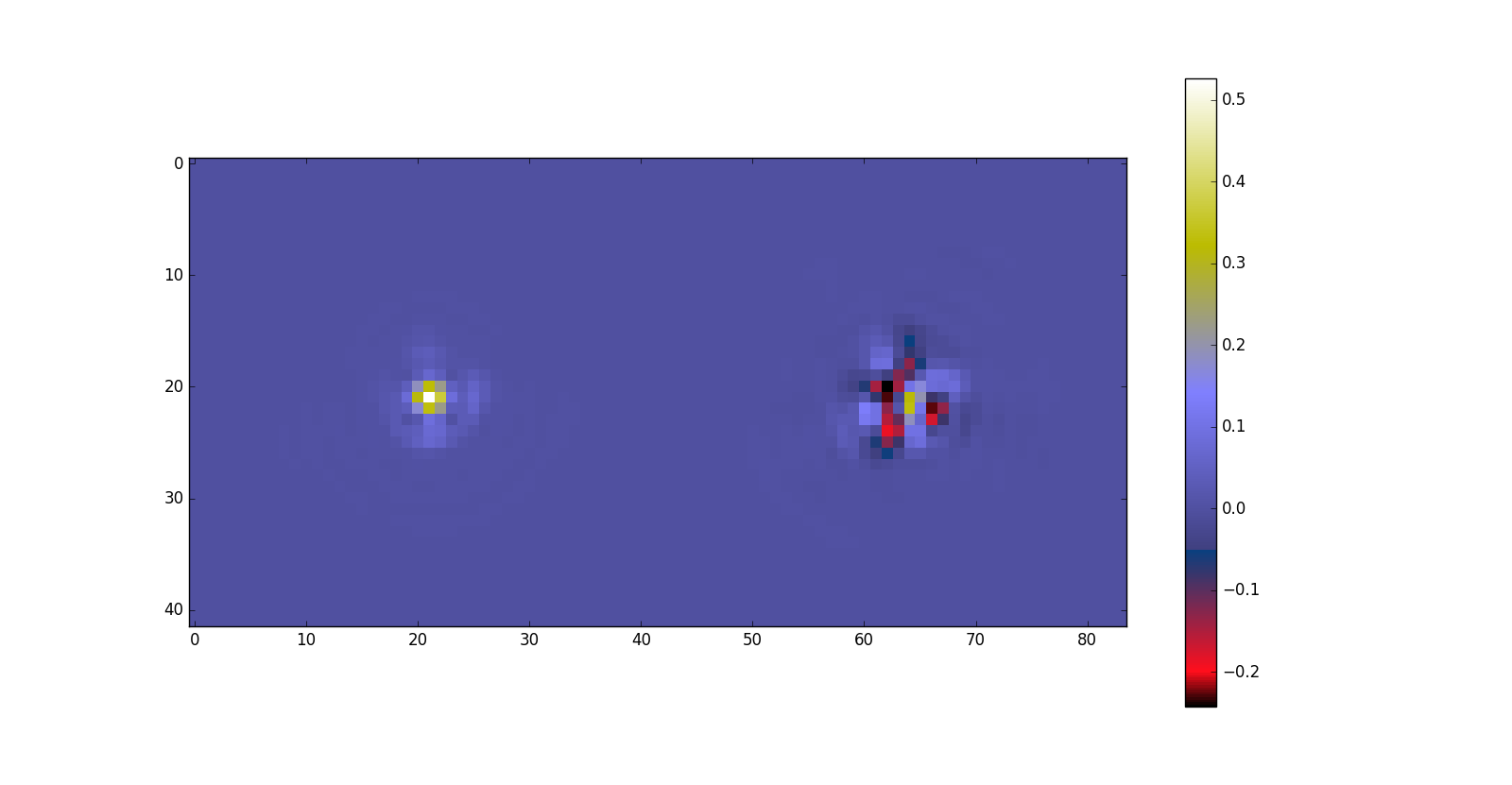}
\end{center}
\caption{First and second principal components; the brightest ring is broken apart between the two components.}
\label{pca_1_2}
\end{figure}

As to the size, it is clearly determined by the brightest structures on the PSF, namely the main lobe and the first brightest ring. The Figure \ref{pca_1_2} shows that the brightest ring information is distributed onto several components, in particular the first and the second one. One expects a bias in the RBF interpolation from processing this information component-wise. On the contrary, the brightest features are globally and more accurately modeled through the barycentric coordinates calculation step, since they influence it the most. Therefore, TraIn gives the most accurate results in terms of size.   

\section{Practical considerations}
\label{practical}
\subsection{Parameters}
\label{param}
\begin{figure}
\begin{center}
\includegraphics[scale=0.6]{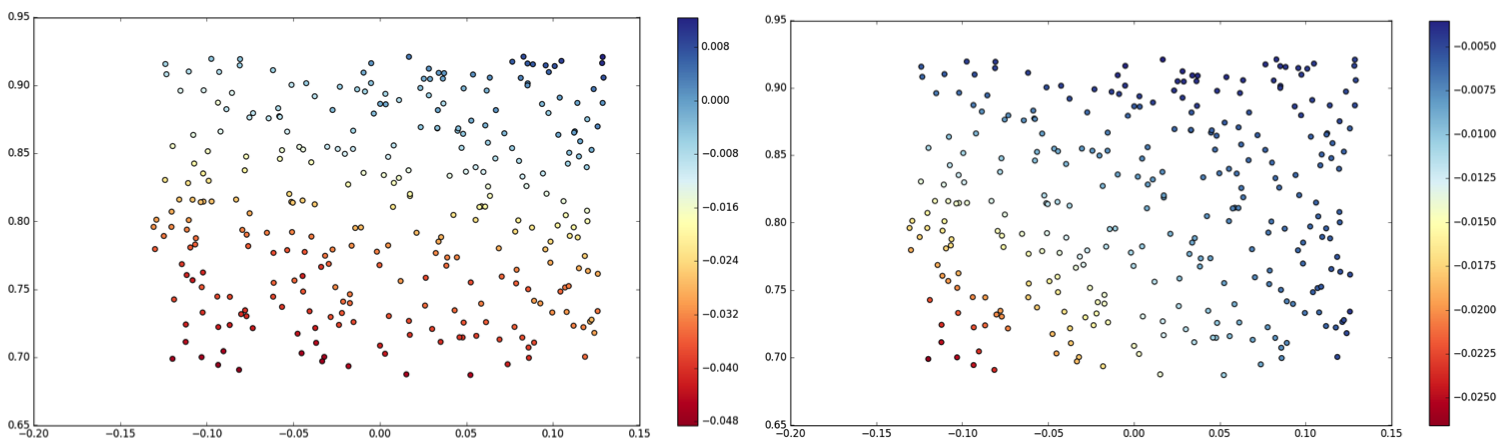}
\end{center}
\caption{Spatial distributions of the first (left) and second ellipticity (right) parameters of the observed PSFs.}
\label{ell_field}
\end{figure}

\paragraph{Local dimensionality} We refer to the parameter $d$ involved in the step 7 in Algorithm \ref{algo}. For an ideally dense data set in the sense of the underlying manifold, this parameter can be set to the manifold intrinsic dimension which is 2 in this example\cite{intrin_dim}; however this is tricky since the ideal density depends on the manifold geometric complexity which is a priori not known. We define the "extrinsic dimensionality" $d_{\text{ext}}$ of the data set as the dimension of the smaller subspace containing the data set. This can be approximately estimated using a PCA. Then we set $d$ to $\text{min}(p,d_{\text{ext}})$, where $p$ is the number of neighbors.    
\paragraph{Number of neighbors} The proposed method gives quite stable results with respect to the number of neighbors. However, Fig.~\ref{ell_field} suggests a potential improvement of the neighbors selection. For a given position of interpolation, the selection of neighbors can be pictured as follows:
\begin{itemize}
\item one initializes a null radius sphere centered on the interpolation position in the FOV;
\item one increases the sphere radius until it includes a given number of observations locations.
\end{itemize}

Yet, Fig.~\ref{ell_field} shows that the ellipticity parameters change faster in certain directions than in others. 
This suggests to grow the neighborhoods according to the ellipticity parameters spatial gradients in such a way to include more neighbors from directions where  the ellipticity parameters vary slowly. 

\paragraph{Ground metric weight parameter} We consider the parameter $\beta$ in Eq.~\ref{gnd}. $\beta$ is the cost of matching two neighbor pixels (assuming a 4-connected neighborhood) with identical intensities. Let consider two images $\mathbf{x} = (x_i)_{1\leq i\leq N}$ and $\mathbf{y} = (y_i)_{1\leq i\leq N}$ and the associated point clouds $\mathbf{X}$ and $\mathbf{Y}$ in the sense of Section \ref{data_rep}. If $\forall i \in \llbracket 1,N \rrbracket |x_i-y_i|\leq \beta$ then $W_2(\mu_{\mathbf{X}},\mu_{\mathbf{Y}}) = \|\mathbf{x}-\mathbf{y}\|_2$. Thus, $\beta$ somehow determines how much the Wasserstein geometry is susceptible to depart from the pixels domain Euclidean geometry. For sufficiently small neighborhoods around a given location in the FOV, the pixels domain Euclidean distance describes accurately the local geometry of the PSFs manifold. Therefore, $\beta$ has to be chosen so that the Wassertein distance is equal to the pixels domain Euclidean distance on any small vicinity in the FOV. Hence the following procedure for setting $\gamma$:
\begin{itemize}
\item determine the two closest PSFs in the FOV $\mathbf{x}_i$ and $\mathbf{x}_j$;
\item set $\gamma$ to $\|\mathbf{x}_i-\mathbf{x}_j\|_{\infty}$.
\end{itemize}
In words, $\gamma$ is chosen as the largest absolute pixel's intensity difference between the two closest observed PSFs in the FOV. 

\subsection{Transportation problems}
\label{tp_solving}
We recall the definition of the sliced Wasserstein distance between two clouds $\mathbf{X}$ and $\mathbf{Y}$:
\begin{equation}
SW_2(\mu_{\mathbf{X}},\mu_{\mathbf{Y}})^2 = \int_{\mathbb{S}^{d-1}} W_2(\mu_{\mathbf{X}_{\mathbf{u}}},\mu_{\mathbf{Y}_{\mathbf{u}}})^2 d\mathbf{u},
\end{equation}
where $\mathbb{S}^{d-1}= \{\mathbf{x}\in \mathbb{R}^d / \|\mathbf{x}\| = 1\}$, $\mathbf{X}_{\mathbf{u}} = \{\mathbf{u}^T\mathbf{x}_i, i = 1...N \} \subset \mathbb{R}^N$ and $\mathbf{Y}_{\mathbf{u}}$ is similarly defined.

\paragraph{Pairwise distances computation: assignments discrepancies} The continuous integration being intractable, in practice, the sliced Wassertein distance is calculated as a discrete sum:
\begin{equation}
SW_2(\mu_{\mathbf{X}},\mu_{\mathbf{Y}})^2 = \sum_{\mathbf{u}_i \in \Omega \subset \mathbb{S}^{d-1}} W_2(\mu_{\mathbf{X}_{\mathbf{u}_i}},\mu_{\mathbf{Y}_{\mathbf{u}_i}})^2.
\end{equation}
For each vector $\mathbf{u}_i$ we recall that the Wassertein 2 distance takes the form
\begin{equation}
W_2(\mu_{\mathbf{X}_{\mathbf{u}_i}},\mu_{\mathbf{Y}_{\mathbf{u}_i}})^2 = \sum_{j=1}^N ((\mathbf{u}_i^T\mathbf{X})[j]-(\mathbf{u}_i^T\mathbf{Y})[\sigma_{\mathbf{u}_i}^*(j)])^2,
\end{equation}
where $\sigma_{\mathbf{u}_i}$ is a permutation of $\llbracket 1,N\rrbracket$. We use the stochastic gradient descent algorithm proposed in \cite{sliced_wass} for calculating $SW_2$. The algorithm estimates a local minimum $\mathbf{Y}^*$ of the functional $J_{\mathbf{Y}}(\mathbf{Z}) = SW_2(\mathbf{Z},\mathbf{Y})^2$ in the vicinity of $\mathbf{X}$ and the approximated Wasserstein distance is computed as 
\begin{equation}
W_2(\mathbf{X},\mathbf{Y}) \approx \|\mathbf{X}-\mathbf{Y}^*\|_2.
\label{wass_approx}
\end{equation}

At each gradient step, the "sliced assignments" $(\sigma_{\mathbf{u}_i}^*)_i$ are updated. To guarantee the convergence, one can use a decreasing step size $\equiv \frac{1}{n^a}$ for $a \in \rrbracket 1/2,1\rrbracket$, $n$ being the iteration index \cite{bottou1998online}.  

The algorithm succeeds in computing an assignment between the clouds $\mathbf{X}$ and $\mathbf{Y}$ if the slices assignments are identical at convergence and if the stationary point $\mathbf{Y}^*$ represents the same point cloud as $\mathbf{Y}$; these two conditions are summarized below:
\begin{gather}
SW_2(\mu_{\mathbf{Y}^*},\mu_{\mathbf{Y}})^2 = \sum_{\mathbf{u}_i \in \Omega \subset \mathbb{S}^{d-1}} \sum_{j=1}^N ((\mathbf{u}_i^T\mathbf{Y}^*)[j]-(\mathbf{u}_i^T\mathbf{Y})[\sigma_{\mathbf{u}_i}^*(j)])^2 = 0 \text{ and}\\
\text{for } \mathbf{u}_i \neq \mathbf{u}_j, \; \sigma_{\mathbf{u}_i}^* = \sigma_{\mathbf{u}_j}^*.
\label{success_cond_1}
\end{gather}
The displacement interpolation between the clouds represented by $\mathbf{X}$ and $\mathbf{Y}$ is realized by performing a linear interpolation between the matrices $\mathbf{X}$ and $\mathbf{Y}^*$. 
We define the discrepancy support as the following set
\begin{equation}
\mathcal{D} = \{k \in \llbracket 1,N\rrbracket / \exists (\mathbf{u}_i,\mathbf{u}_j) \in \Omega^2 /\sigma_{\mathbf{u}_i}^*[k] \neq \sigma_{\mathbf{u}_j}^*[k]\}
\label{disc_supp}
\end{equation}
and define the assignment discrepancy as the support size $|\mathcal{D}|$.

As observed in \cite{sliced_wass} and \cite{sliced_bonneel}, we did not find it necessary in practice to use a decreasing step size. We observe as well that the stationary points seem to always satisfy $|\mathcal{D}|=0$. However, it is not rare that the sequence of iterates oscillates around a point which has a non-zero discrepancy support size in which case, the algorithm takes a potentially long time to reach a stationary point satisfying $|\mathcal{D}|=0$.

\begin{figure}
\begin{center}
\includegraphics[scale=0.8]{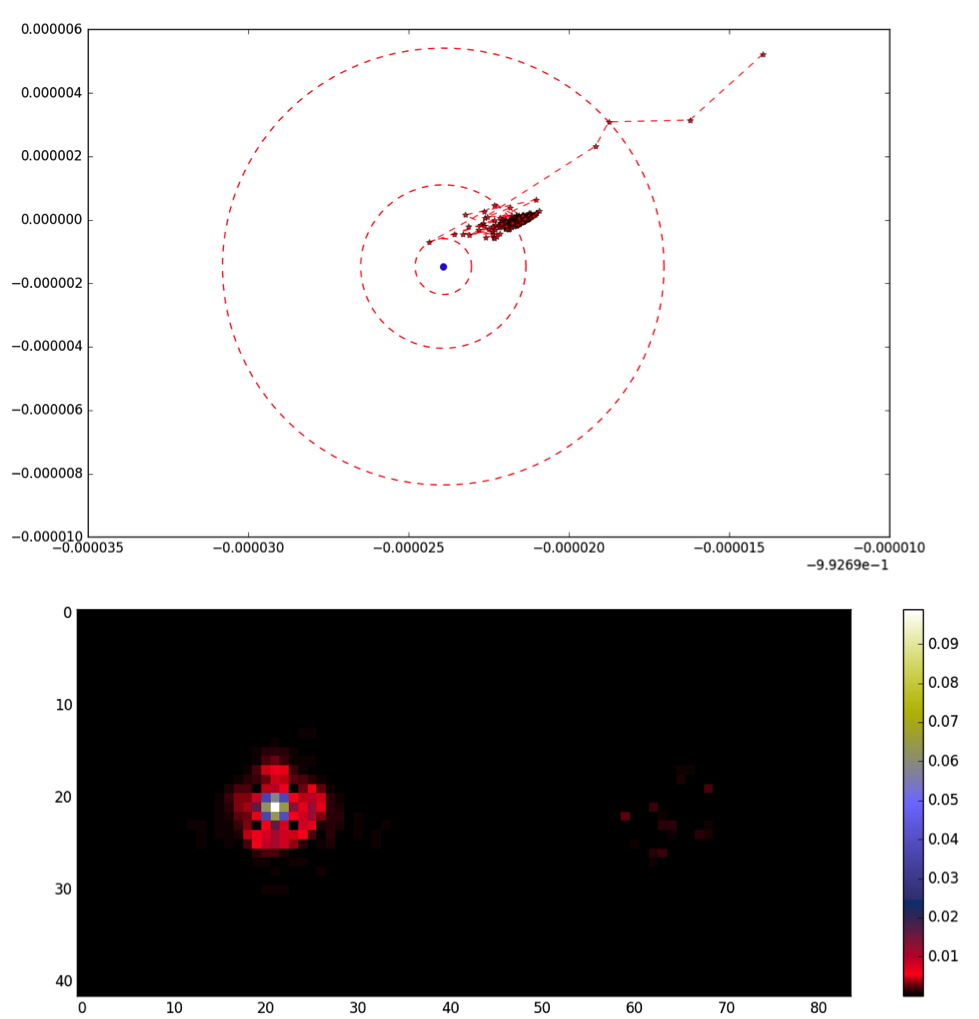}
\end{center}
\caption{Assignment failure: on the top plot, the sequence of images associated with each iterate is represented by the red dots and the image associated with the ideal solution is represented by the blue dot; these images are shown in the plan spanned by the two first principal components calculated from the whole images set; the iterates end up oscillating around a point on the middle circle; the bottom left plot shows the ideal image and the bottom right plot shows the absolute difference between the final image and the ideal one; one can see that there are some non-negligible artifacts.}
\label{assign_fail}
\end{figure}

In case $|\mathcal{D}|>0$ for the final iterate $\mathbf{Y}^*$, it mixes information from different pixels in the image associated to $\mathbf{Y}$. This translates into visual artifacts when $\mathbf{Y}^*$ is transformed into an image (see \ref{im_point_cloud}). One can see such an example in Fig.~\ref{assign_fail}. This is a troublesome point since in the considered astronomical application, the systematic errors due to the PSFs estimation constitute one of the bottlenecks.
 
\begin{figure}
\begin{center}
\includegraphics[scale=0.7]{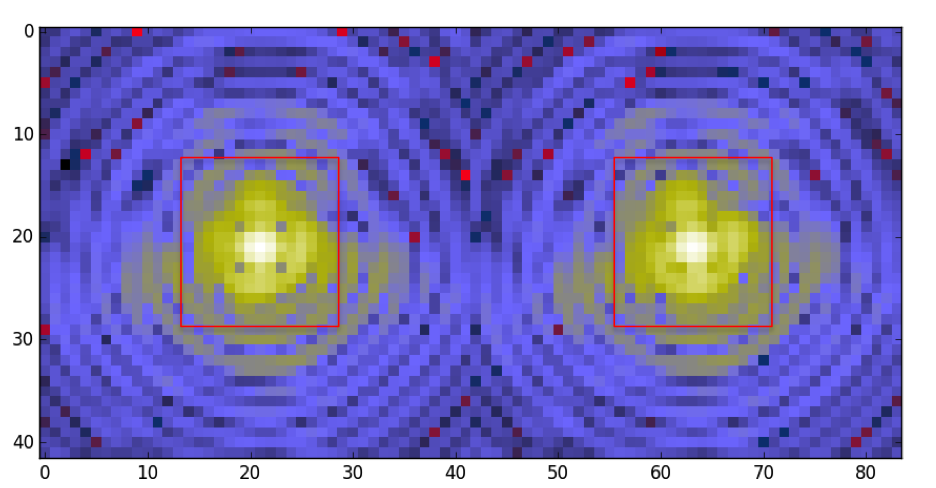}
\end{center}
\caption{Two PSFs examples in a logarithmic scale; we first use the hungarian algorithm to map the two point clouds derived from the pixels located in the small red rectangles in each PSF respectively; the result is used for initializing the sliced transport algorithm on the point clouds associated with the full images (see \ref{im_mod}).}
\label{crop}
\end{figure}

Besides, the algorithm is stochastic and the energy minimized is non-convex; therefore for the same input point clouds, the algorithm converges to a different point at each run. We propose to reduce the final assignment discrepancy with a fixed number of iterations and the stationary point variability by improving the gradient descent initialization. Taking the two point clouds $\mathbf{X}$ and $\mathbf{Y}$, we extract from each cloud the points corresponding the pixels comprised in a small rectangular window (typically of size $20\times 20$) around the corresponding images centroids (see Fig.~\ref{crop}).
Let note $\textbf{ind}_X$ and $\textbf{ind}_Y$ the sets of corresponding indexes and  $\overline{\textbf{ind}}_{X}$ the complement of $\textbf{ind}_X$ in $\llbracket 1,N\rrbracket$. Then we find the optimal assignment between $\mathbf{X}[\textbf{ind}_X]$ and $\mathbf{Y}[\textbf{ind}_Y]$ using the Hungarian algorithm\cite{kuhn50hungarian}, which is the fastest known procedure for solving exactly the assignment problem. However, it has a roughly cubic complexity in its improved version, which restricts its practical use to small scale problems.  We note $\sigma$ the optimal mapping:
\begin{equation}
\mathbf{X}[\textbf{ind}_X[i]] \mapsto \mathbf{Y}[\textbf{ind}_Y[\sigma(i)]], \forall i \in \llbracket 1,|\textbf{ind}_X|\rrbracket.
\end{equation}
Then we initialize the aforementioned stochastic gradient descent method with the matrix $\mathbf{X}_0$ defined as follows:

\begin{equation}
\left\lbrace \begin{matrix}\mathbf{X}_0[:,i] = \mathbf{X}[:,i] \text{ if } i \in \overline{\textbf{ind}}_{X}\text{ and } \\
\mathbf{X}_0[:,\textbf{ind}_X[i]] = \mathbf{Y}[\textbf{ind}_Y[\sigma(i)]] \text{ for } i \in \llbracket 1,|\textbf{ind}_X|\rrbracket.
\label{def_init} 
\end{matrix}\right.
\end{equation}

\begin{figure}
\begin{center}
\includegraphics[scale=0.55]{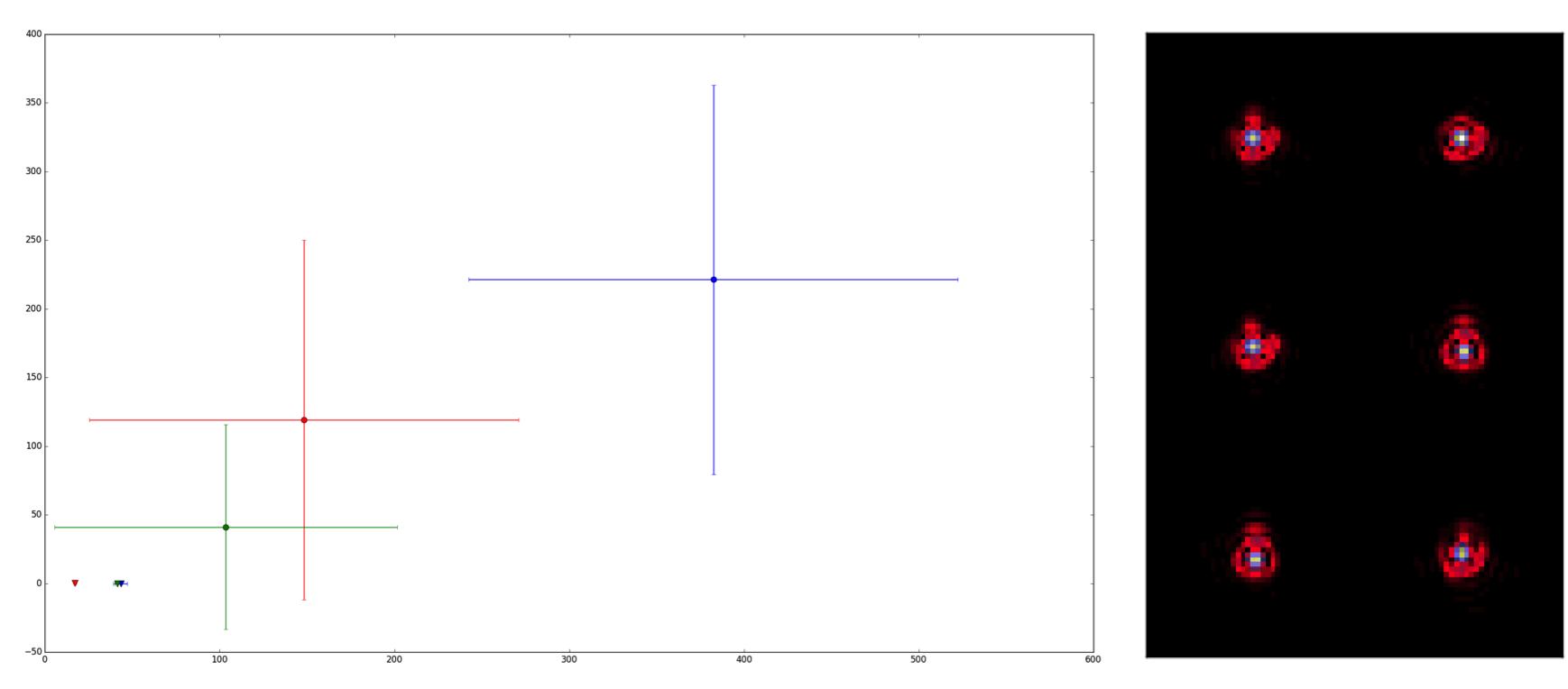}
\end{center}
\caption{Sliced Wasserstein transport algorithm average performances; x axis: average running times in seconds; y axis: average final discrepancy support size; the red, green and yellow correspond to the pairs of images from the top to the bottom respectively; the triangle-shaped points correspond to the gradient descents with an optimized initialization; the points are obtained by averaging these performances criteria over 100 runs for the 3 pairs of $42 \times 42$ images; the proposed initialization dramatically reduces the time needed to compute an assignment between our images related point clouds with an average execution time ranging from 10 to 40s rather than 100 to 400s.}
\label{smart_init}
\end{figure}

To quantify the impact of this pre-assignment, we compare the average running time of the algorithm, for 3 different pairs of images, for 100 runs in each case, with and without optimizing the initialization. These times account for the hungarian algorithm when used. We set a maximum number of iterations of 20000. The discrepancy support size $|\mathcal{D}|$ might be non-zero when this number of iterations is reached. Therefore, we also compute the average of the final discrepancy support sizes. The number of projection directions was set to 30 and the initialization window size to $20\times 20$. As for all the experiments presented, we used $42\times 42$ PSFs images, which gives clouds of $1724$ points each one to compare. The result is displayed in Fig.~\ref{smart_init}.  

This results follows from the simple observation that the higher is $|\mathcal{D}|$ for a given iterate, the more the stochastic gradient descent is oscillatory. The Figure \ref{smart_init} shows that it is possible to set the exact matching window size so that the Hungarian algorithm brings more benefits in terms of discrepancies reduction than it increases the computational complexity. 

It is worth noting that the assignment obtained from the sliced Wasserstein transport depends on the number of slices directions. As outlined in \cite{sliced_bonneel}, the more directions there are, the smoother is the assignment. Unsurprisingly, this dependency is reduced when involving the Hungarian algorithm. Actually in that case, we obtain exactly the same assignment independently of the number of directions, for two given PSFs images and an initialization window of size $20\times 20$. 

\begin{figure}
\begin{center}
\includegraphics[scale=0.85]{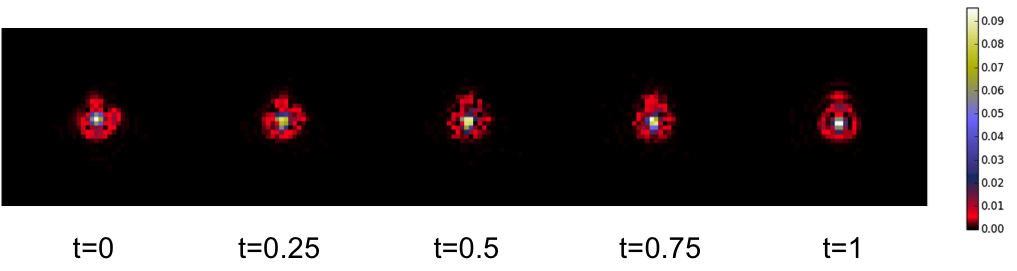}
\end{center}
\caption{Displacement interpolation: "shrinkage effect"; the interpolated PSFs have a narrower energy distribution than the initial and final ones.}
\label{disp_interp_naive}
\end{figure}

\begin{figure}
\begin{center}
\includegraphics[scale=0.81]{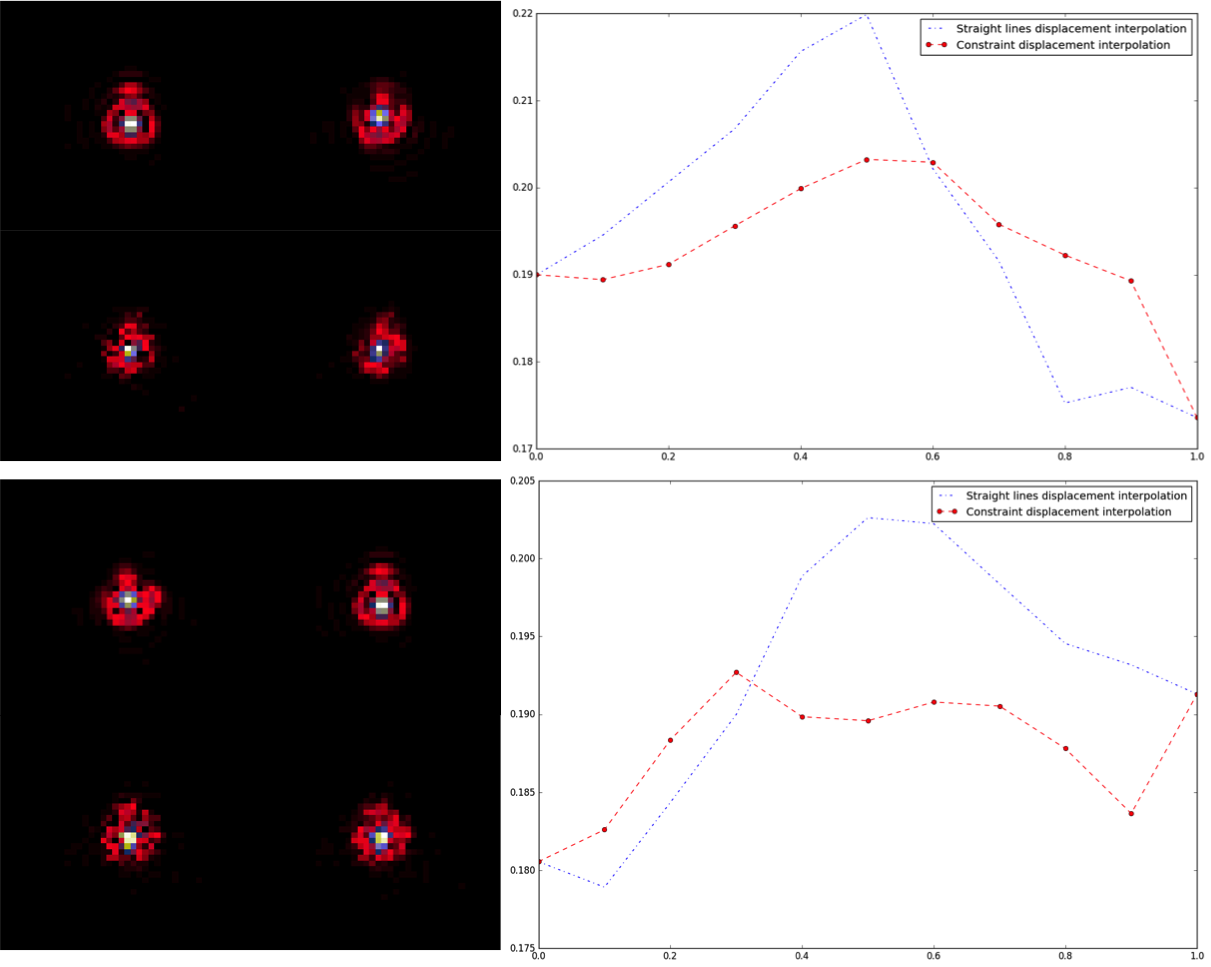}
\end{center}
\caption{Constraint displacement interpolation: the two PSFs panels show original PSFs on the top and the interpolated PSFs for t=0.5 are shown on the bottom left for the regular displacement interpolation and the bottom right for the velocity constraint displacement interpolation; the right plots show the evolution of the $l_2/l_1$ norms ratio of interpolated PSFs as functions of the time parameter in the two examples respectively; the velocity constraint reduces the shrinkage effect since the corresponding $l_2/l_1$ norms ratios are flattened compared to those obtained with the regular displacement interpolation; moreover, the interpolated images are smoother.}
\label{disp_interp_mod}
\end{figure}

\paragraph{Velocity constraint displacement interpolation}
The Wasserstein barycenter of two point clouds is simply calculated by realizing a displacement interpolation (see Section \ref{not}). Considering 2 PSFs, a simplified interpolation process in this framework would go as follows:
\begin{itemize}
\item the PSFs are first converted into two matrices $\mathbf{X}$ and $\mathbf{Y}$ representing two different point clouds;
\item the point clouds are matched according to the previous paragraph; this yields a matrix $\mathbf{Y}^*$ representing the same point cloud as $\mathbf{Y}$;
\item the point clouds interpolation is realized by moving each point in the first point cloud toward the matched point in the second point cloud along a straight line and at a uniform speed on the cloud; this is done by performing a linear interpolation of the matrices $\mathbf{X}$ and $\mathbf{Y}^*$;
\item the interpolated PSFs are obtained by converting interpolated clouds at different "times" into images.  
\end{itemize}
An example is shown in Fig.~\ref{disp_interp_naive}. In this example, the brightest PSF's ring is more elongated horizontally for the initial PSF and vertically for the final PSF. But as the displacement interpolation constraints particles to move along straight lines, the interpolated point clouds are shrunken compared to the extreme ones and consequently, the interpolated images have a more concentrated energy. Quantitatively, the interpolated PSFs tend to have $l_2/l_1$  norms ratio that significantly exceed the range observed over the data set - note that the PSFs $l_1$ norms are constant and equal to 1.    
This problem can be approached from the ground metric angle; precisely, one can think of choosing a ground metric that admits curved geodesics, making a more "compact" particles advection possible. However, the complex variation of the PSFs across the FOV makes the choice of a better fitted ground metric non trivial. 
The displacement interpolation generates a sequence of matrices $(\mathbf{X}_i)_{1\leq i\leq T}$ so that $\mathbf{X}_1 = \mathbf{X}$, $\mathbf{X}_T = \mathbf{Y}^*$ and for $i \in \llbracket 1,T-1\rrbracket$, $\mathbf{X}_{i+1}-\mathbf{X}_i = \mu_i(\mathbf{Y}^*-\mathbf{X})$   
for a step $\mu_i$.  
In order to better preserve the point clouds densities in the displacement interpolation, we impose the points to have \underline{parallel velocity vectors} at each time. To do so, we modify the displacement interpolation as follows:
\begin{enumerate}
\label{const_disp_interp}
\item INPUT: a maximum step size $\mu_{\text{max}}$
\item Initialization: $\mathbf{X}_0 = \mathbf{X}, i=0$.
\item WHILE $i\leq i_{\text{max}}$: \\
	$\> \mathbf{V}_i = \mathbf{Y}^*-\mathbf{X}_i$\\
	$\>$ Compute an eigenvector $\mathbf{u}_i$ corresponding to the highest eigenvalue of $\mathbf{V}_i\mathbf{V}_i^T$\\
	$\> \widehat{\mathbf{V}}_i = \mathbf{u}_i\mathbf{u}_i^T \mathbf{V}_i/\|\mathbf{u}_i\|_2^2$\\
	$\>$ Line search: $\mu_{\text{opt}} = \underset{\mu}\argmin \|\mathbf{X}_i+\mu\widehat{\mathbf{V}}_i - \mathbf{Y}^*\|_2^2 =- \frac{\langle \widehat{\mathbf{V}}_i,\mathbf{X}_i-\mathbf{Y}^*\rangle}{\|\widehat{\mathbf{V}}_i\|_2^2}$\\
	$\> \mathbf{X}_{i+1} = \mathbf{X}_i+\text{min}(\mu_{\text{opt}},\mu_{\text{max}})\widehat{\mathbf{V}}_i$\\
 	$\>i = i+1$.
\end{enumerate}
In the above procedure, all the particles are moved in parallel with $\mathbf{u}_i$ at each iteration. Moreover, $\|\mathbf{X}_{i+1} - \mathbf{Y}^*\|_2^2 \leq \|\mathbf{X}_i - \mathbf{Y}^*\|_2^2$. Therefore the procedure moves in fact the initial point cloud toward the final one. We do not discuss how close the sequence generated can come to $\mathbf{Y}^*$. However, in the tests we performed, the distance $\|\mathbf{X}_i-\mathbf{Y}^*\|$ always gets down to the machine numerical precision.
In Fig.~\ref{disp_interp_naive}, the time parameter is defined as 
\begin{equation}
t(i) = 1 - \frac{\|\mathbf{X}_i-\mathbf{Y}^*\|_2}{\|\mathbf{X}-\mathbf{Y}^*\|_2},
\label{time_param_1}
\end{equation}
which makes sense in a dynamic model of constant speed advection along straight lines. However, in the modified displacement interpolation, the matrices $(\mathbf{X}_i)_{1\leq i\leq i_{\text{max}}}$ do not necessarily follow a linear path in the matrix space. To make a direct comparison possible between the two displacement interpolations, the time parameter definition has to be generalized in order to account for a possible curvature of the path. We do so by involving the length of the curve obtained by joining each matrix of the sequence to the following one with a straight line:
\begin{equation}
t(i) = 1 - \frac{\sum_{j=i}^{i_{\text{max}}}\|\mathbf{X}_i-\mathbf{X}_{i+1}\|_2}{\sum_{j=0}^{i_{\text{max}}}\|\mathbf{X}_i-\mathbf{X}_{i+1}\|_2},
\label{time_param_gen}
\end{equation}
with the convention $\mathbf{X}_{i_{\text{max}}+1} = \mathbf{Y}^*$. This definition generalizes \ref{time_param_1}.
We compare the two displacement interpolations in Fig.~\ref{disp_interp_mod}. The parallel velocity constraint quantitatively reduces the shrinkage effect since the interpolated images $l_2/l_1$ norms ratios do not exceed those of the reference images as much as with the regular displacement interpolation. Not surprisingly, this constraint yields smoother barycenters.

This constraint can be integrated directly into the sliced OT algorithm. However, we observe that this considerably increases the number of iterations needed to converge and most importantly, the generated sequence mostly converges to a solution which is not a global minimum of the functional $J_{\mathbf{Y}}(\mathbf{Z}) = SW_2(\mathbf{Z},\mathbf{Y})^2$.
We note that this constraint can be naturally integrated into the framework adopted in \cite{not_prox} since velocities are directly manipulated. However, the eulerian discretization would require manipulating 4 dimensional arrays, making the approach intractable. 

This constraint has not been used for generating the plots shown in Section \ref{num_res}. Indeed, for that experiment, we considered a density of known PSFs that makes the PSFs variations in a given neighborhood in the FOV very smooth in which case the shrinkage effect disappears. 

\paragraph{Wasserstein barycenters computation} As explained in Section \ref{train_meth}, the interpolated PSF's computation involves calculating a sliced Wassertein barycenter. Let consider a set of point clouds $(\mathbf{Y}_i)_{1\leq i\leq p}$. As for the pairwise Wasserstein distances, an approximation of their Wasserstein barycenter is approximately calculated by finding a local stationary point of the functional $J_{\mathbf{Y}_1,\dots,\mathbf{Y}_p}(\mathbf{Z}) = \sum_{i=1}^p w_i SW_2(\mathbf{Z},\mathbf{Y}_i)^2$\cite{sliced_wass}, for some barycentric weights $(w_i)_{1\leq i\leq p}$. The higher is $p$, the more degenerated is this functional because of its combinatorial nature, which increases the assignments discrepancies.  But as mentioned in Section \ref{wass_bar_section}, there is no efficient method for calculating exactly a Wasserstein barycenter of more than two point clouds, even for small scale problems. Therefore, the previously described strategy for speeding up the convergence can not be directly extended to this case. We propose to approximate the Wasserstein barycenter of a set of more than two point clouds by computing a sequence of "$2$ points" Wasserstein barycenters. This relies on the local isometry assumption made in Section \ref{embedding}. Indeed, the barycenter of a set of vectors $(\mathbf{r}_i)_{1\leq i\leq p}$ associated with the barycentric weights $(w_i)_{1\leq i\leq p}$ can be calculated through the following procedure:

\begin{enumerate}
\label{partial_barycenter}
\item Initialization: $\mathbf{r}_{\text{bar}} = \mathbf{r}_1; \; i=2,\; w = w_1 $
\item WHILE $i\leq p$: \\
  $\> \mathbf{r}_{\text{bar}} = \underset{\mathbf{r}}\argmin\; w_i \|\mathbf{r}- \mathbf{r}_i\|_2^2+ w\|\mathbf{r}- \mathbf{r}_{\text{bar}}\|_2^2 $\\
 $\> w = w+w_i$\\
 $\>i = i+1$.
\end{enumerate}
Indeed, one can check that at the $i^{\text{th}}$ iteration $\mathbf{r}_{\text{bar}}$ is updated to $\sum_{k=1}^i \frac{w_k}{\sum_{j=1}^i w_j} \mathbf{r}_k$.
If the Wasserstein space is isometric to a Euclidean space in the neighborhood of the point clouds involved in the barycenter's calculation, then one can apply a similar procedure in the Wasserstein space to calculate the barycenter, hence the following scheme:

\begin{enumerate}
\label{partial_barycenter_wass}
\item Initialization: $\mathbf{Z}_{\text{bar}} = \mathbf{Z}_1; \; i=2,\; w = w_1 $
\item WHILE $i\leq p$: \\
  $\> \mathbf{Z}_{\text{bar}} = \underset{\mathbf{Z}}\argmin\; w_i SW_2(\mathbf{Z},\mathbf{Y}_i)^2+ w SW_2(\mathbf{Z},\mathbf{Z}_{\text{bar}})^2 $\\
 $\> w = w+w_i$\\
 $\>i = i+1$.
\end{enumerate}
This way we can take advantage of the previous accurate initializing proposed in the Wasserstein barycenter approximation and make use of the $1D$ displacement interpolation. 
The computed barycenter slightly changes depending on the ordering of the clouds $\mathbf{Y}_i$. We choose the ordering so that $w_1\geq \dots \geq w_p$.

\section{Reproducible research}
In the spirit of participating in reproducible research, the data and the codes used to generate the plots presented in this paper will be made available at \url{http://www.cosmostat.org/software/}.

\section{Conclusion} 
We introduce TraIn (Transport Interpolation) which is a data field interpolation method based on Optimal Transport and making use of some Manifold Learning ideas. We consider the interpolation of a PSF field. The local geometry of the PSF field is characterized using approximated Wasserstein distances. From these, we derive a low dimensional local Euclidean embedding of the PSF field which is then mapped to the instrument field-of-view using a thin-plate interpolation. This mapping gives one the embedded coordinates of the unknown PSFs, from which a representation of these PSFs in the Wasserstein metric is determined. Finally, the interpolated PSFs are calculated through nested displacement interpolations.

We compared TraIn to the Inverse Distance Weighting method and to a component-wise thin-plate interpolation of PCA coefficients. The tests were made on a set of realistic Euclid-like\cite{Eucl1} PSFs. We show that the proposed method is globally the most accurate in terms of pixel domain error and shape. In particular, TraIn is in average several orders of magnitudes more accurate than the two other methods in terms of pixels errors when a few number of "neighbor PSFs" are used for the interpolation.
We also introduce a velocity constraint displacement interpolation for mitigating the unnatural shape shrinkage that might occur when interpolating objects with different major axis using an Euclidean ground metric.

A natural extension of this study would be to compare these interpolation methods in a setting where the reference PSFs are not perfectly known.

\section{Acknowledgments}
This work is supported by  
the European Community through the grants PHySIS  (contract no. 640174) and  DEDALE  (contract no. 665044) within the H2020 Framework Program. The authors acknowledge the Euclid Collaboration, the European Space Agency and the support of the Centre National d’Etudes Spatiales.
The authors wish to thank Koryo Okumura, Patrick Hudelot and Jérôme Amiaux for providing Euclid-like PSFs.

\bibliographystyle{siam}
\bibliography{ref}

\begin{thebibliography}{10}

\bibitem{Abdi2007}
{\sc H.~Abdi}, {\em Metric multidimensional scaling}, in Encyclopedia of
  Measurement and Statistics., N.J. Salkind, ed., Sage, Thousand Oaks (CA),
  2007, pp.~598--605.

\bibitem{sliced_bonneel}
{\sc N.~Bonneel, J.~Rabin, G.~Peyr{\'e}, and H.~Pfister}, {\em Sliced and radon
  wasserstein barycenters of measures}, Journal of Mathematical Imaging and
  Vision, 51 (2015), pp.~22--45.

\bibitem{bottou1998online}
{\sc L.~Bottou}, {\em Online learning and stochastic approximations}, On-line
  learning in neural networks, 17 (1998), p.~142.

\bibitem{opt_assign}
{\sc R.~Burkard, M.~Dell'Amico, and S.~Martello}, {\em Assignment Problems},
  Society for Industrial and Applied Mathematics, Philadelphia, PA, USA, 2009.

\bibitem{OEOF}
{\sc P.~Clarysse, B.~Delhay, M.~Picq, and J.~Pousin}, {\em Optimal extended
  optical flow subject to a statistical constraint}, Journal of Computational
  and Applied Mathematics, 234 (2010), pp.~1291 -- 1302.

\bibitem{dense_flow_est}
{\sc T.~Corpetti, E.~Memin, and P.~Perez}, {\em Dense estimation of fluid
  flows}, IEEE Transactions on Pattern Analysis and Machine Intelligence, 24
  (2002), pp.~365--380.

\bibitem{Crop1}
{\sc M.~Cropper~et al.}, {\em {Defining a weak lensing experiment in space}},
  Monthly Notices of the Royal Astronomical Society, 431 (2013).

\bibitem{cuturi_1}
{\sc M.~Cuturi and A.~Doucet}, {\em Fast computation of wasserstein
  barycenters}, in Proceedings of the 31st International Conference on Machine
  Learning (ICML-14), 2014, pp.~685--693.

\bibitem{dokmanic2015Euclidean}
{\sc I.~Dokmanic, R.~Parhizkar, J.~Ranieri, and M.~Vetterli}, {\em Euclidean
  distance matrices: Essential theory, algorithms, and applications}, IEEE
  Signal Processing Magazine, 32 (2015), pp.~12--30.

\bibitem{Donoho03hessianeigenmaps}
{\sc D.~L. Donoho and C.~Grimes}, {\em Hessian eigenmaps: New locally linear
  embedding techniques for high-dimensional data}, 2003.

\bibitem{eberly2002thin}
{\sc D.~Eberly}, {\em Thin plate splines}, Geometric Tools Inc., 2002.,
  (2002).

\bibitem{epperson1987runge}
{\sc J.~F. Epperson}, {\em On the runge example}, American Mathematical
  Monthly, 94 (1987), pp.~329--341.

\bibitem{Eucl1}
{\sc ESA/SRE}, {\em {EUCLID Mapping the geometry of the dark universe}}, tech.
  report, ESA, July 2011.

\bibitem{fortun2015optical}
{\sc D.~Fortun, P.~Bouthemy, and C.~Kervrann}, {\em Optical flow modeling and
  computation: a survey}, Computer Vision and Image Understanding, 134 (2015),
  pp.~1--21.

\bibitem{gangbo}
{\sc W.~Gangbo and A.~Swiech}, {\em Optimal maps for the multidimensional
  monge-kantorovich problem}, Communications on Pure and Applied Mathematics,
  51 (1998), pp.~23--45.

\bibitem{gentile_2013}
{\sc M.~{Gentile}, F.~{Courbin}, and G.~{Meylan}}, {\em {Interpolating point
  spread function anisotropy}}, Astronomy \& Astrophysics, 549 (2013), p.~A1.

\bibitem{goodman2005introduction}
{\sc J.~W. Goodman}, {\em Introduction to Fourier optics}, Roberts and Company
  Publishers, 2005.

\bibitem{gower}
{\sc J.~C. Gower}, {\em Adding a point to vector diagrams in multivariate
  analysis}, Biometrika, 55 (1968), pp.~582--585.

\bibitem{gramf}
{\sc A.~Gramfort, G.~Peyr{\'e}, and M.~Cuturi}, {\em Fast optimal transport
  averaging of neuroimaging data}, in International Conference on Information
  Processing in Medical Imaging, Springer, 2015, pp.~261--272.

\bibitem{haker2004optimal}
{\sc S.~Haker, L.~Zhu, A.~Tannenbaum, and S.~Angenent}, {\em Optimal mass
  transport for registration and warping}, International Journal of computer
  vision, 60 (2004), pp.~225--240.

\bibitem{jarv_2004}
{\sc M.~{Jarvis} and B.~{Jain}}, {\em {Principal Component Analysis of PSF
  Variation in Weak Lensing Surveys}}, ArXiv Astrophysics e-prints,  (2004).

\bibitem{jar_2008}
{\sc M.~{Jarvis}, P.~{Schechter}, and B.~{Jain}}, {\em {Telescope Optics and
  Weak Lensing: PSF Patterns due to Low Order Aberrations}}, ArXiv e-prints,
  (2008).

\bibitem{jee_2007}
{\sc M.~J. Jee, J.~P. Blakeslee, M.~Sirianni, A.~R. Martel, R.~L. White, and
  H.~C. Ford}, {\em Principal component analysis of the time- and
  position-dependent point-spread function of the advanced camera for surveys},
  Publications of the Astronomical Society of the Pacific, 119 (2007), pp.~pp.
  1403--1419.

\bibitem{kaiser}
{\sc N.~{Kaiser}, G.~{Squires}, and T.~{Broadhurst}}, {\em {A Method for Weak
  Lensing Observations}}, The Astrophysical Journal, 449 (1995), p.~460.

\bibitem{kant}
{\sc L.~V. Kantorovich}, {\em {On the translocation of masses}}, Compte-Rendu
  de l'Académie des Sciences de l'USSR, 321 (1942), pp.~199--201.

\bibitem{fire_detect}
{\sc I.~Kolesov, P.~Karasev, A.~Tannenbaum, and E.~Haber}, {\em Fire and smoke
  detection in video with optimal mass transport based optical flow and neural
  networks}, in 2010 IEEE International Conference on Image Processing, Sept
  2010, pp.~761--764.

\bibitem{kuhn50hungarian}
{\sc H.~W. Kuhn}, {\em The hungarian method for the assignment problem}, 50
  Years of Integer Programming 1958--2008, p.~29.

\bibitem{intrin_dim}
{\sc A.V. Little, J.~Lee, J.~Yoon-Mo, and M.~Maggioni}, {\em Estimation of
  intrinsic dimensionality of samples from noisy low-dimensional manifolds in
  high dimensions with multiscale svd}, in Statistical Signal Processing, 2009.
  SSP '09. IEEE/SP 15th Workshop on, Aug 2009, pp.~85--88.

\bibitem{mds}
{\sc J.~I. Marden}, {\em Analyzing and modeling rank data}, London ; New York :
  Chapman \& Hall, 1st ed.~ed., 1995.

\bibitem{shap2}
{\sc R.~{Massey} and A.~{Refregier}}, {\em {Polar shapelets}}, Monthly Notices
  of the Royal Astronomical Society, 363 (2005), pp.~197--210.

\bibitem{McCann1997153}
{\sc R.~J. McCann}, {\em A convexity principle for interacting gases}, Advances
  in Mathematics, 128 (1997), pp.~153 -- 179.

\bibitem{mueller2013optical}
{\sc M.~Mueller, P.~Karasev, I.~Kolesov, and A.~Tannenbaum}, {\em Optical flow
  estimation for flame detection in videos}, IEEE Transactions on image
  processing, 22 (2013), pp.~2786--2797.

\bibitem{fng}
{\sc F~Ngolè, J-L Starck, K~Okumura, J~Amiaux, and P~Hudelot}, {\em Constraint
  matrix factorization for space variant psfs field restoration}, Inverse
  Problems, 32 (2016), p.~124001.

\bibitem{not_prox}
{\sc N.~Papadakis, G.~Peyré, and E.~Oudet}, {\em Optimal transport with
  proximal splitting}, SIAM Journal on Imaging Sciences, 7 (2014),
  pp.~212--238.

\bibitem{piot}
{\sc L.~W. {Piotrowski}, T.~{Batsch}, H.~{Czyrkowski}, M.~{Cwiok},
  R.~{Dabrowski}, G.~{Kasprowicz}, A.~{Majcher}, A.~{Majczyna}, K.~{Malek},
  L.~{Mankiewicz}, K.~{Nawrocki}, R.~{Opiela}, M.~{Siudek}, M.~{Sokolowski},
  R.~{Wawrzaszek}, G.~{Wrochna}, M.~{Zaremba}, and A.~F. {{\.Z}arnecki}}, {\em
  {PSF modelling for very wide-field CCD astronomy}}, Astronomy \&
  Astrophysics, 551 (2013), p.~A119.

\bibitem{sliced_wass}
{\sc J.~Rabin, G.~Peyré, J.~Delon, and M.~Bernot}, {\em Wasserstein barycenter
  and its application to texture mixing.}, in Scale Space and Variational
  Methods in Computer Vision, Alfred~M. Bruckstein, Bart~M. ter Haar~Romeny,
  Alexander~M. Bronstein, and Michael~M. Bronstein, eds., vol.~6667 of Lecture
  Notes in Computer Science, Springer, 2011, pp.~435--446.

\bibitem{shap1}
{\sc A.~{Refregier}}, {\em {Shapelets - I. A method for image analysis}},
  Monthly Notices of the Royal Astronomical Society, 338 (2003), pp.~35--47.

\bibitem{romano}
{\sc A.~{Romano}, L.~{Fu}, F.~{Giordano}, R.~{Maoli}, P.~{Martini},
  M.~{Radovich}, R.~{Scaramella}, V.~{Antonuccio-Delogu}, A.~{Donnarumma},
  S.~{Ettori}, K.~{Kuijken}, M.~{Meneghetti}, L.~{Moscardini},
  S.~{Paulin-Henriksson}, E.~{Giallongo}, R.~{Ragazzoni}, A.~{Baruffolo},
  A.~{Dipaola}, E.~{Diolaiti}, J.~{Farinato}, A.~{Fontana}, S.~{Gallozzi},
  A.~{Grazian}, J.~{Hill}, F.~{Pedichini}, R.~{Speziali}, R.~{Smareglia}, and
  V.~{Testa}}, {\em {Abell 611. I. Weak lensing analysis with LBC}}, Astronomy
  \& Astrophysics, 514 (2010), p.~A88.

\bibitem{stab}
{\sc H.~F. {Stabenau}, B.~{Jain}, G.~{Bernstein}, and M.~{Lampton}}, {\em
  {Lensing Systematics from Space: Modeling PSF effects in the SNAP survey}},
  ArXiv e-prints,  (2007).

\bibitem{interp_cs}
{\sc A.~B. {Suksmono}}, {\em {Reconstruction of Complex-Valued Fractional
  Brownian Motion Fields Based on Compressive Sampling and Its Application to
  PSF Interpolation in Weak Lensing Survey}}, ArXiv e-prints,  (2013).

\bibitem{tenenbaum_global_2000}
{\sc J.~B. Tenenbaum, V.~de~Silva, and J.~C. Langford}, {\em A global geometric
  framework for nonlinear dimensionality reduction}, Science, 290 (2000),
  p.~2319.

\bibitem{vill_2}
{\sc Cédric Villani}, {\em Optimal transport : old and new}, Grundlehren der
  mathematischen Wissenschaften, Springer, Berlin, 2009.

\end{thebibliography}

\Appendix
\section{Multidimensional scaling}
\label{mds_cent}
We consider a family of vectors $(\mathbf{r}_i)_{1\leq i\leq p}$ in $\mathbb{R}^q$; we set $\mathbf{R} = [\mathbf{r}_1,\dots,\mathbf{r}_p]$. We assume that $\sum_{i=1}^p \mathbf{r}_i = 0_{\mathbb{R}^q}$. We define the distances matrix $\mathbf{D} = (\mathbf{\|\mathbf{r}_i-\mathbf{r}_j\|_2^2})_{1\leq i,j\leq p}$ and the norms vector $\mathbf{v} = (\|\mathbf{r}_1\|_2^2,\dots,\|\mathbf{r}_p\|_2^2)^T$. We want to retrieve $(\mathbf{r}_i)_{1\leq i\leq p}$ from $\mathbf{D}$. 
$\mathbf{D}$ can be rewritten as:
\begin{equation}
\mathbf{D} = \mathbf{v}\mathds{1}_p^T+\mathds{1}_p\mathbf{v}^T-2\mathbf{R}^T\mathbf{R}.
\label{dist_mat}
\end{equation}
We consider the centering matrix $\mathbf{C}$ introduced in Section \ref{embedding}. We have $\mathbf{C}\mathds{1}\mathbf{v}^T = 0_{\mathbb{R}^p} = \mathbf{v}\mathds{1}_p^T\mathbf{C}$. Thus $-\frac{1}{2}\mathbf{C}\mathbf{D}\mathbf{C} = (\mathbf{R}\mathbf{C})^T\mathbf{R}\mathbf{C}$. But $\mathbf{R}\mathbf{C} = (\sum_{i=1}^p \mathbf{r}_i)\mathds{1}_p^T=0$, which finally gives $-\frac{1}{2}\mathbf{C}\mathbf{D}\mathbf{C} = \mathbf{R}^T\mathbf{R}$. 

\section{point clouds to image transform}
\label{pt_cloud_image}
\begin{figure}
\begin{center}
\includegraphics[scale=0.5]{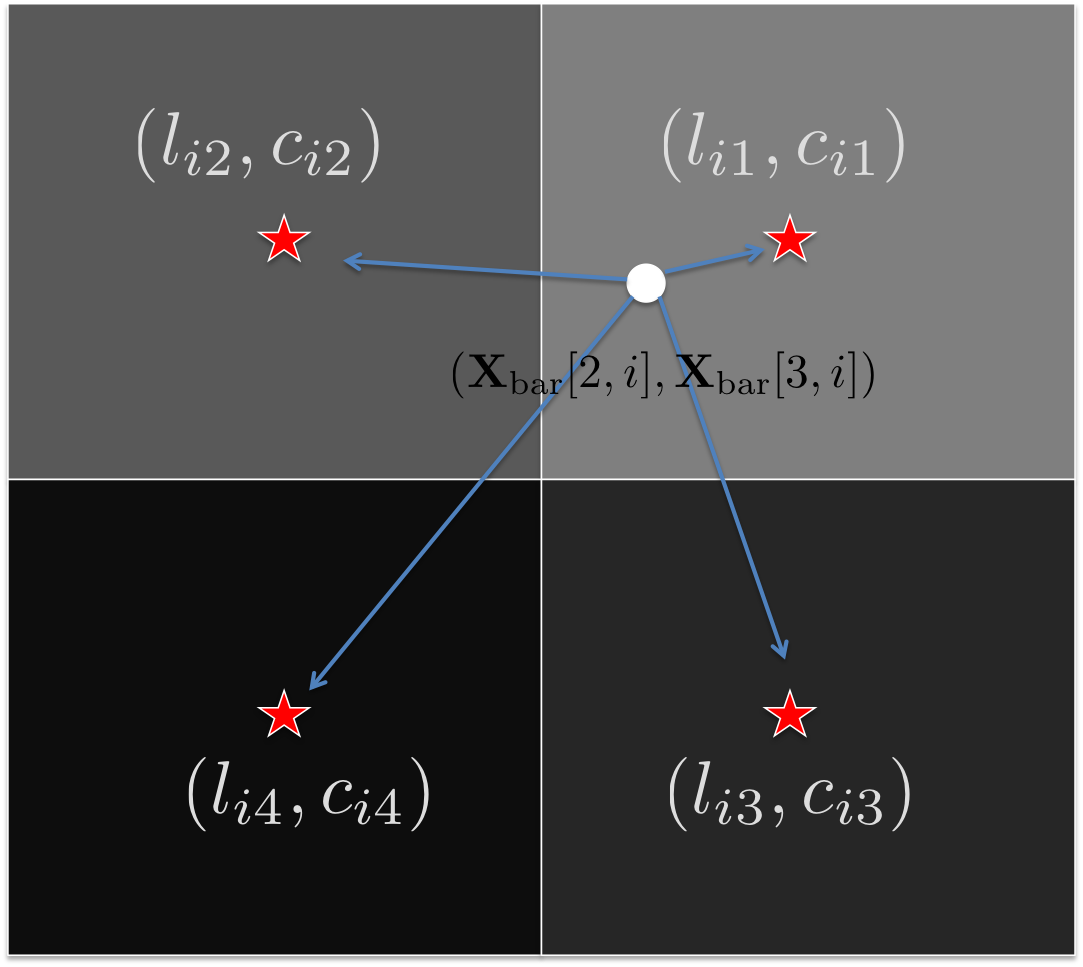}
\end{center}
\caption{Point cloud to image mapping.}
\label{loc_cloud_to_pix}
\end{figure}

The step 11 in the Algorithm \ref{algo} consists in converting estimated Wasserstein barycenters which are point clouds into images which are the actual PSFs estimates. Let consider an estimated Wasserstein $\mathbf{X}_{\text{bar}} \in \mathbb{R}^{3 \times N}$. By construction, the first dimension is related to pixel intensities and the second and third are related to pixel positions (see Eq.~\ref{im_mod}). We assume that the PSFs images have $N_l$ lines and $N_c$ columns so that $N = N_l N_c$. Then the most simple way of calculating an image from $\mathbf{X}_{\text{bar}}$ is the following:
\begin{enumerate}
\item Initialize an $N_l\times N_c$ image $\mathbf{\text{PSF}}_{\text{bar}}$ with all pixels values set to zeros.
\item FOR ALL $i\in \llbracket 1,N\rrbracket$:\\
$\> $ find the nearest neighbor $(l_i,c_i)$ of $\mathbf{X}_{\text{bar}}[2:3,i]$ in $\llbracket 1,N_l\rrbracket\times \llbracket 1,N_c\rrbracket.$\\
$\> $ Set $\mathbf{\text{PSF}}_{\text{bar}}[l_i,c_i] = \mathbf{\text{PSF}}_{\text{bar}}[l_i,c_i]+\mathbf{X}_{\text{bar}}[1,i].$ 
\end{enumerate}
However, this generally results in sharp pixel intensities variations and therefore visual artifacts.  We circumvent this by involving the four nearest pixels in step 2 in the procedure above, rather than the nearest one solely:  
\begin{enumerate}
\item Initialize an $N_l\times N_c$ image $\mathbf{\text{PSF}}_{\text{bar}}$ with all pixels values set to zeros.
\item FOR ALL $i\in \llbracket 1,N\rrbracket$:\\
$\> $ find the 4 nearest neighbors $((l_{ij},c_{ij}))_{1\leq j \leq 4}$ of $\mathbf{X}_{\text{bar}}[2:3,i]$ in $\llbracket 1,N_l\rrbracket\times \llbracket 1,N_c\rrbracket.$\\
$\>$ FOR ALL $j \in \llbracket 1,4\rrbracket$:\\
$\>$ set $\mathbf{\text{PSF}}_{\text{bar}}[l_{ij},c_{ij}] = \mathbf{\text{PSF}}_{\text{bar}}[l_{ij},c_{ij}]+\frac{1}{\sum_{k=1}^4 \frac{\|\mathbf{X}_{\text{bar}}[2:3,i]-(l_{ij},c_{ij})^T\|_2^2}{\|\mathbf{X}_{\text{bar}}[2:3,i]-(l_{ik},c_{ik})^T\|_2^2}}\mathbf{X}_{\text{bar}}[1,i].$ 
\end{enumerate}
This is illustrated in Fig.~\ref{loc_cloud_to_pix}. 

\section{Ellipticity parameters directional derivatives}
\label{ell_der}
The ellipticity parameters \ref{ellipticity} can be rewritten in the following way:
\begin{gather}
e_1(\mathbf{X}_i)= \frac{<\mathbf{X}_i,\mathbf{U}_5><\mathbf{X}_i,\mathbf{U}_3>-<\mathbf{X}_i,\mathbf{U}_1>^2+<\mathbf{X}_i,\mathbf{U}_2>^2}{<\mathbf{X}_i,\mathbf{U}_4><\mathbf{X}_i,\mathbf{U}_3>-<\mathbf{X}_i,\mathbf{U}_1>^2-<\mathbf{X}_i,\mathbf{U}_2>^2} \\
e_2(\mathbf{X}_i)= \frac{2(<\mathbf{X}_i,\mathbf{U}_6><\mathbf{X}_i,\mathbf{U}_3>-<\mathbf{X}_i,\mathbf{U}_1><\mathbf{X}_i,\mathbf{U}_2>)}{<\mathbf{X}_i,\mathbf{U}_4><\mathbf{X}_i,\mathbf{U}_3>-<\mathbf{X}_i,\mathbf{U}_1>^2-<\mathbf{X}_i,\mathbf{U}_2>^2},
\label{ellipticity_2}
\end{gather} 
where $\mathbf{U}_1 = (k)_{1\leq k\leq N_l \atop 1\leq l\leq N_c},\; \mathbf{U}_2 = (l)_{1\leq k\leq N_l \atop 1\leq l\leq N_c},\; \mathbf{U}_3 = (1)_{1\leq k\leq N_l \atop 1\leq l\leq N_c},\; \mathbf{U}_4 = (k^2+l^2)_{1\leq k\leq N_l \atop 1\leq k\leq N_c},\; \mathbf{U}_5 = (k^2-l^2)_{1\leq k\leq N_l \atop 1\leq l\leq N_c},\; \mathbf{U}_6 = (kl)_{1\leq k\leq N_l \atop 1\leq l\leq N_c}$.
We derive the following expressions:
\begin{gather}
\frac{de_1(\mathbf{X}_i+t\mathbf{P}_j)}{dt} = \frac{a_1+a_2 t}{c_t} - e_1(\mathbf{X}_i+t\mathbf{P}_j) \frac{d_1+d_2 t}{c_t}\\
\frac{de_2(\mathbf{X}_i+t\mathbf{P}_j)}{dt} = \frac{b_1+b_2 t}{c_t} - e_2(\mathbf{X}_i+t\mathbf{P}_j) \frac{d_1+d_2 t}{c_t},
\label{ell_deriv}
\end{gather}
where 
\begin{itemize}
\item $c_t =  <\mathbf{U}_4, \mathbf{X}_i+t\mathbf{P}_j><\mathbf{U}_3, \mathbf{X}_i+t\mathbf{P}_j> - <\mathbf{U}_1, \mathbf{X}_i+t\mathbf{P}_j>^2 -<\mathbf{U}_2, \mathbf{X}_i+t\mathbf{P}_j>^2$
\item $a_1 = <\mathbf{U}_5,\mathbf{X}_i><\mathbf{U}_3,\mathbf{P}_j>+ <\mathbf{U}_3,\mathbf{X}_i><\mathbf{U}_5,\mathbf{P}_j>-2(<\mathbf{U}_1,\mathbf{X}_i><\mathbf{U}_1,\mathbf{P}_j>-<\mathbf{U}_2,\mathbf{X}_i><\mathbf{U}_2,\mathbf{P}_j>)$
\item $a_2 = 2(<\mathbf{U}_5, \mathbf{P}_j><\mathbf{U}_3, \mathbf{P}_j> - <\mathbf{U}_1, \mathbf{P}_j>^2 +<\mathbf{U}_2,\mathbf{P}_j>^2)$
\item $b_1 = 2(<\mathbf{U}_6,\mathbf{X}_i><\mathbf{U}_3,\mathbf{P}_j>+<\mathbf{U}_3,\mathbf{X}_i><\mathbf{U}_6,\mathbf{P}_j>-<\mathbf{U}_1,\mathbf{X}_i><\mathbf{U}_2,\mathbf{P}_j>-<\mathbf{U}_2,\mathbf{X}_i><\mathbf{U}_1,\mathbf{P}_j>)$
\item $b_2 = 4(<\mathbf{U}_6,\mathbf{P}_j><\mathbf{U}_3,\mathbf{P}_j>-<\mathbf{U}_2,\mathbf{P}_j><\mathbf{U}_1,\mathbf{P}_j>)$
\item $d_1 = <\mathbf{U}_4,\mathbf{X}_i><\mathbf{U}_3,\mathbf{P}_j>+ <\mathbf{U}_3,\mathbf{X}_i><\mathbf{U}_4,\mathbf{P}_j>-2(<\mathbf{U}_1,\mathbf{X}_i><\mathbf{U}_1,\mathbf{P}_j>+<\mathbf{U}_2,\mathbf{X}_i><\mathbf{U}_2,\mathbf{P}_j>)$
\item $d_2 = 2(<\mathbf{U}_4, \mathbf{P}_j><\mathbf{U}_3, \mathbf{P}_j> - <\mathbf{U}_1, \mathbf{P}_j>^2 -<\mathbf{U}_2,\mathbf{P}_j>^2)$.
\end{itemize}

%
%
%
%

\end{document}